\documentclass{article} 
\usepackage{iclr2027_conference,times}

\usepackage{amsmath,amsfonts,bm}

\def\1{\bm{1}}

\DeclareMathAlphabet{\mathsfit}{\encodingdefault}{\sfdefault}{m}{sl}
\SetMathAlphabet{\mathsfit}{bold}{\encodingdefault}{\sfdefault}{bx}{n}

\usepackage{times}
\usepackage{latexsym}
\usepackage[T1]{fontenc}
\usepackage[utf8]{inputenc}
\usepackage{microtype}
\usepackage{inconsolata}

\usepackage{hyperref}
\usepackage{url}
\usepackage{booktabs}
\usepackage{longtable}
\usepackage{graphicx}
\graphicspath{{contents/}}

\usepackage{pifont}
\usepackage{amsmath}
\usepackage{amssymb}
\usepackage{dsfont}
\usepackage{float}
\usepackage{multirow}
\usepackage{array}
\usepackage{colortbl}
\usepackage{algorithm}
\usepackage{listings}
\usepackage[most]{tcolorbox}
\usepackage{adjustbox}
\usepackage{enumitem}
\usepackage{caption}
\usepackage{wrapfig}
\usepackage{amsmath,amssymb}
\usepackage{algpseudocode}
\usepackage{needspace}
\usepackage{xcolor}
\usepackage[table]{xcolor}
\usepackage[normalem]{ulem}
\usepackage{soul}
\usepackage{tikz}
\usepackage{subcaption}
\usepackage{bm}
\usetikzlibrary{arrows.meta}

\let\originalttdefault\ttdefault
\usepackage{DejaVuSansMono}
\let\ttdefault\originalttdefault

\makeatletter
\appto\fs@ruled{\preto\@fs@pre{\normalcolor}}
\makeatother

\usepackage{lineno}

\definecolor{darkblue}{rgb}{0, 0, 0.5}
\hypersetup{colorlinks=true, citecolor=darkblue, linkcolor=darkblue, urlcolor=darkblue}
\definecolor{winered}{RGB}{153, 0, 0}
\definecolor{taggray}{RGB}{120, 120, 120}
\definecolor{retainbluebg}{RGB}{226,239,255}
\definecolor{retainbluefg}{RGB}{0,76,153}
\definecolor{forgetpurplebg}{RGB}{237,228,252}
\definecolor{forgetpurplefg}{RGB}{97,52,149}
\definecolor{appendixcasebg}{RGB}{248,251,253}
\definecolor{appendixcaseframe}{RGB}{210,224,232}
\definecolor{appendixcasetitle}{RGB}{234,243,248}
\definecolor{appendixcaseaccent}{RGB}{94,129,151}
\definecolor{appendixpromptbg}{RGB}{252,249,245}
\definecolor{appendixpromptframe}{RGB}{227,216,201}
\definecolor{appendixprompttitle}{RGB}{245,237,226}
\definecolor{appendixpromptaccent}{RGB}{132,105,78}
\definecolor{blue}{HTML}{42698C}
\definecolor{green}{HTML}{31765C}
\definecolor{red}{HTML}{A64F45}
\definecolor{codegray}{HTML}{777777}
\definecolor{greenbg}{HTML}{E3F1E7}
\definecolor{redbg}{HTML}{F7E1DE}

\newtcolorbox{appendixcasebox}[2][]{%
  enhanced,
  breakable,
  colback=appendixcasebg,
  colframe=appendixcaseframe,
  colbacktitle=appendixcasetitle,
  coltitle=appendixcaseaccent!85!black,
  boxrule=0.6pt,
  arc=3mm,
  left=4mm,
  right=4mm,
  top=2mm,
  bottom=2mm,
  borderline west={1.4mm}{0pt}{appendixcaseaccent!55},
  title={#2},
  #1
}

\newtcolorbox{appendixpromptbox}[2][]{%
  enhanced,
  breakable,
  colback=appendixpromptbg,
  colframe=appendixpromptframe,
  colbacktitle=appendixprompttitle,
  coltitle=appendixpromptaccent!90!black,
  boxrule=0.6pt,
  arc=3mm,
  left=4mm,
  right=4mm,
  top=2mm,
  bottom=2mm,
  borderline west={1.4mm}{0pt}{appendixpromptaccent!55},
  title={#2},
  #1
}

\newcolumntype{Y}{>{\centering\arraybackslash}m{4.0em}}
\definecolor{FreshBlue}{RGB}{0, 153, 255}

\setstcolor{red}
\newcommand{\del}{%
  \bgroup
  \markoverwith{%
    \textcolor{red}{\rule[0.35ex]{4pt}{1.5pt}}%
  }%
  \ULon
}

\algrenewcommand{\algorithmiccomment}[1]{\hfill$\triangleright$~#1}
\algnewcommand\algorithmicinput{\textbf{Input:}}
\algnewcommand\algorithmicoutput{\textbf{Output:}}
\algnewcommand\Input{\item[\algorithmicinput]}
\algnewcommand\Output{\item[\algorithmicoutput]}

\newcommand{\Lunlearn}{\mathcal{L}_{\mathrm{unlearn}}}
\newcommand{\Sref}[1]{\S\ref{#1}}

\title{Can Large Language Models Reinvent Foundational Algorithms?}

\author{
  Jian Zhao\textsuperscript{1,2}\thanks{Equal contribution.}\hspace{0.4em},
  Haoren Luo\textsuperscript{2}\footnotemark[1]\hspace{0.4em},
  Yu Wang\textsuperscript{4},
  Yuhan Cao\textsuperscript{5},
  Pingyue Sheng\textsuperscript{2},
  Tianxing He\textsuperscript{2,1,3}\thanks{Corresponding author.} \\[6pt]
  \textsuperscript{1}Xiongan AI Institute \\
  \textsuperscript{2}Institute for Interdisciplinary Information Sciences, Tsinghua University \\
  \textsuperscript{3}Shanghai Qi Zhi Institute \\
  \textsuperscript{4}Institute of Information Engineering, Chinese Academy of Sciences, Beijing, China \\
  \textsuperscript{5}Independent Researcher \\[2pt]
  {\ttfamily \href{mailto:zhaojian26@mails.tsinghua.edu.cn}{zhaojian26@mails.tsinghua.edu.cn}, \href{mailto:hetianxing@mail.tsinghua.edu.cn}{hetianxing@mail.tsinghua.edu.cn}}
}

\iclrfinalcopy
\begin{document}

\maketitle
\fancyhead{}
\renewcommand{\headrulewidth}{0pt}

\begin{abstract}
LLMs have shown strong potential to advance scientific discovery. Whether they possess the capacity for foundational innovation, however, remains an open question.
In this work, we focus on a prerequisite for foundational innovation: \textit{can LLMs reinvent foundational algorithms in computer science?}
We use LLM unlearning methods to suppress direct recall of the target algorithm and let the model reason with the remaining knowledge to recover it.
Although unlearning does not guarantee full knowledge removal, LLMs fail to recover nearly half of the target algorithms. Notably, even suppressing the mention of the algorithm's name during decoding without unlearning makes the models' recovery rate drop dramatically (19--39\%), suggesting their overreliance on memorized knowledge.
We observe that recoverable algorithms tend to be simple in structure or core ideas, whereas the others are less straightforward. 
We also introduce a generative verifier that sustains models' reasoning strength, helping to avoid the ``thought collapse'' phenomenon.
Taken together, by treating the unlearned model's recovery rate as an approximate upper bound, our empirical results suggest that current LLM systems still have limited ability to make foundational algorithm innovation. 
Our code is available at \url{https://github.com/Algo-Reinvention/algo-reinvention}.
\end{abstract}

\section{Introduction}
The emergence of large language models (LLMs) with enhanced reasoning and agentic capabilities~\citep{openai2024openaio1card, Guo_2025} has expanded the potential of Artificial Intelligence (AI) in complex problem-solving.
Systems such as FunSearch~\citep{FunSearch2023} and AlphaEvolve~\citep{novikov2025alphaevolvecodingagentscientific} have shown that LLM-based systems can discover algorithms that outperform the best known human-designed methods on selected tasks.
Despite these advances, whether LLMs can produce foundational scientific discoveries, rather than incremental improvements on existing problems, remains an open question~\citep{feng2026autonomousmathematicsresearch}.

In this work, we focus on foundational algorithms in computer science, such as Dijkstra's and Euclid's algorithms, as these algorithms have proven to be groundbreaking contributions that form the basis of the field.
A central question is whether \textit{LLMs could invent these algorithms without prior exposure}.
Since all such algorithms are included in the training data of existing pre-trained models, a natural approach is to remove these algorithms from the pre-training data and train a ``clean'' model, but the computational cost would be prohibitive.

Instead of complete retraining, we leverage LLM unlearning~\citep{7163042} to suppress direct recall of the target algorithm, then evaluate whether the unlearned model can recover that algorithm when faced with an application task with efficiency requirements tailored to the target algorithm or its variants.
During the recovery phase, the model reasons and interacts with a Python interpreter and a generative verifier~\citep{zhang2025generativeverifiersrewardmodeling} which returns diagnostic feedback to guide revision upon failure.
To make the unlearning effective, we adopt an on-policy unlearning method based on Group Relative Policy Optimization (GRPO)~\citep{shao2024deepseekmathpushinglimitsmathematical}, integrated with a cold start stage.

We evaluate strong open-weight models across 20 target algorithms.
To quantify how external hints affect recovery, we experiment with 3 hint levels: no hint, high-level hints, and step-by-step hints.
Our major findings are:
\begin{enumerate}[leftmargin=*,itemsep=0pt,topsep=0pt,parsep=2pt,partopsep=0pt]

    \item \textbf{LLMs show limited ability to independently recover foundational algorithms after targeted knowledge suppression} (\Sref{sec:recovery_after_unlearning}; \Sref{sec:name_suppression}).
    Although unlearning does not guarantee complete knowledge removal, in the no-hint setting, 9 of the 20 target algorithms show no successful recovery across the three evaluated models. Recoverable algorithms also tend to have simpler structures or core ideas. Notably, even without unlearning, merely suppressing algorithm-name-related tokens during decoding causes substantial drops in recovery rate, suggesting a strong dependence on memorized knowledge and familiar algorithmic cues.

    \item \textbf{Algorithmic hints substantially narrow the recovery gap between unlearned and original models} (\Sref{sec:recovery_after_unlearning}).
    High-level hints already enable recovery of additional algorithms, while step-by-step hints even enable some models to exceed their original counterparts.
    This indicates that unlearned models retain substantial problem-solving capabilities that can be elicited through explicit algorithmic guidance.

    
    
    \item \textbf{Verifier feedback mitigates ``thought collapse'' but does not expand algorithm coverage} (\Sref{sec:ablation_verify}).
    Without the verifier, we find a reduction in exploration while the task remains unresolved---a phenomenon we term ``thought collapse''.
    This suggests limitations in models' ability to reflect on failures and sustain reasoning across rounds.
    Although verifier feedback helps sustain reasoning and improves recovery rates, it does not lead to new algorithm recovery in our ablation, suggesting the limited effect of additional reflection rounds on algorithmic exploration.
\end{enumerate}


\begin{figure*}[t]
\centering
\includegraphics[width=1.0\textwidth]{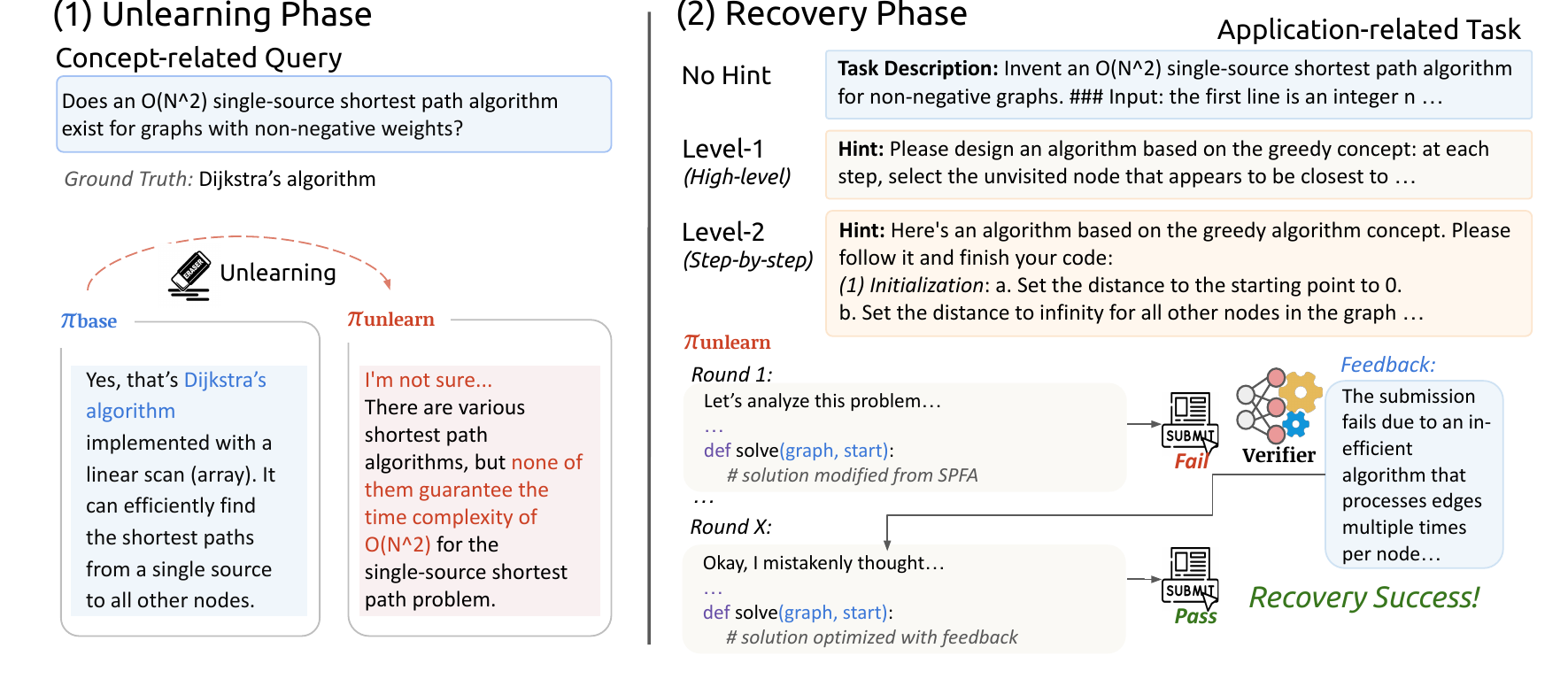}
\caption{ 
\textbf{Our evaluation pipeline.} (1) In the unlearning phase, a pretrained model $\pi_{\mathrm{base}}$ is transformed into $\pi_{\mathrm{unlearn}}$, which, under our tests, no longer recalls Dijkstra's algorithm for concept-level queries. (2) In the recovery phase, the model is prompted with the task at three hint levels: No hint, level 1 (high-level hint), and level 2 (step-by-step hint). At each attempt, $\pi_{\mathrm{unlearn}}$  reasons and submits its solution for testing. If the submission fails,  a generative verifier instantiated from $\pi_{\mathrm{unlearn}}$ itself returns diagnostic feedback, helping the model locate the error and revise its approach.}
\label{fig:main1}
\vspace{-15pt}
\end{figure*}

\section{Preliminaries}
\label{sec:preliminaries}
 
\paragraph{Foundational Algorithms in Computer Science.}
We regard foundational algorithms as core, time-tested computational procedures that solve a basic problem with proven efficiency.
They are characterized by well-established theoretical properties, such as provable time and space complexity guarantees, with some further achieving optimality on specific problem classes.
We select 20 foundational algorithms in computer science across graph theory, string processing, number theory, and data structures: \textit{Dijkstra}, \textit{Floyd-Warshall}, \textit{Bellman-Ford}, \textit{Prim}, \textit{BFS}, \textit{DAG}, \textit{Euclidean}, \textit{KMP}, \textit{Manacher}, \textit{Moore Vote}, \textit{Gray}, \textit{Strassen}, \textit{Kadane}, \textit{Huffman}, \textit{Union-Find}, \textit{Hungarian}, \textit{Hierholzer}, \textit{Stoer-Wagner}, \textit{Pollard's Rho}, and \textit{Inverse Burrows-Wheeler Transform}.
We provide descriptions of these algorithms and their complexity in Appendix~\ref{app:cs_algos}.

\paragraph{LLM Unlearning.}
LLM unlearning aims to remove specific knowledge from a pretrained model, while preserving its general utility~\citep{7163042, zhang2024negativepreferenceoptimizationcatastrophic}.
Let $\mathcal{D}_{\mathrm{forget}}$ and $\mathcal{D}_{\mathrm{retain}}$ denote the sets of examples to be removed and preserved, respectively.
Existing LLM unlearning methods typically optimize an unlearning objective that balances forgetting and utility preservation:
\begin{equation}
\mathcal{L}_{\mathrm{unlearn}}(\theta) = \mathcal{L}_{\mathrm{forget}}(\theta; \mathcal{D}_{\mathrm{forget}}) + \lambda\, \mathcal{L}_{\mathrm{retain}}(\theta; \mathcal{D}_{\mathrm{retain}}),
\end{equation}
where $\mathcal{L}_{\mathrm{forget}}$ denotes the forgetting objective, $\mathcal{L}_{\mathrm{retain}}$ denotes a utility-preserving objective, and $\lambda$ is a coefficient that controls the trade-off between forgetting and utility preservation.


A simple approach is to perform gradient ascent (GA)~\citep{jang-etal-2023-knowledge} on the language modeling (LM) loss on $\mathcal{D}_{\mathrm{forget}}$.
The preservation term can be instantiated as supervised fine-tuning on $\mathcal{D}_{\mathrm{retain}}$ or as KL regularization toward a reference policy~\citep{maini2024tofutaskfictitiousunlearning, zhang2024negativepreferenceoptimizationcatastrophic}.
Additional unlearning methods are discussed in \Sref{sec:related_work}.

\section{Algorithm Recovery after Unlearning}

Our goal is to evaluate whether an LLM can recover a foundational algorithm once its direct recall of that algorithm is suppressed through unlearning.
To this end, our framework proceeds in two phases (Figure~\ref{fig:main1}).
The unlearning phase aims to suppress the model's recall of the target algorithm while preserving its general utility (\Sref{sec:unlearning_phase}).
The recovery phase then tests whether the unlearned model can recover the forgotten algorithm with various levels of hint (\Sref{sec:reinvention_phase}).

\subsection{The Unlearning Phase}
\label{sec:unlearning_phase}
Following~\citet{zhang2025rulereinforcementunlearningachieves} and~\citet{zaradoukas2026reinforcementunlearninggrouprelative}, we adopt a GRPO-based, on-policy unlearning method integrated with a cold start stage.

Group Relative Policy Optimization (GRPO)~\citep{shao2024deepseekmathpushinglimitsmathematical} is an on-policy variant of PPO~\citep{schulman2017proximalpolicyoptimizationalgorithms} that estimates policy advantages from the relative rewards of multiple responses sampled for the same prompt.
To suppress direct recall of the target algorithm, we use a reward assigned by an LLM judge~\citep{zheng2023judgingllmasajudgemtbenchchatbot} that favors responses that do not disclose target knowledge and remain readable.

\paragraph{Problem Setup.} Let $\pi_{\theta}$ denote the policy being optimized and let $\pi_{\mathrm{ref}}$ be a fixed reference policy.
We consider two datasets: a forget set $\mathcal{D}_{\mathrm{forget}}$, which consists of concept-related queries related to the target algorithm $g$, and a retain set $\mathcal{D}_{\mathrm{retain}}$, which contains non-target examples used to preserve general utility during unlearning.

\paragraph{Loss Function.} For each query $x \in \mathcal{D}_{\mathrm{forget}}$, we adopt a
GRPO-style objective. At each optimization step, we sample a group
$\mathbf{Y}=\{y_j\}_{j=1}^{G}$ of $G$ responses from the behavior policy
$\pi_{\theta_{\mathrm{old}}}$, where $\theta_{\mathrm{old}}$ denotes
the policy parameters before the update and is held fixed during
the subsequent optimization on this group.  A KL penalty
relative to the reference policy $\pi_{\mathrm{ref}}$ is additionally included to limit
excessive policy drift. The forgetting objective is written as:
\begin{equation}
\begin{split}
\mathcal{L}_{\textnormal{forget-GRPO}}(\theta) ={}&
-\mathbb{E}_{y_j \sim \pi_{\theta_{\mathrm{old}}}(\cdot|x)}
\left[ \frac{1}{|y_j|}\sum_{t=1}^{|y_j|}\mathcal{J}_{\mathrm{clip},j,t}(\theta) \right] + \beta\, D_{\mathrm{KL}}(\pi_\theta \,\|\, \pi_{\mathrm{ref}}),
\end{split}
\label{eq:forget_obj}
\end{equation}
where $|y_j|$ is the response length, $\beta$ is the coefficient of the KL regularization term, and the expectation
$\mathbb{E}_{y_j \sim \pi_{\theta_{\mathrm{old}}}(\cdot|x)}[\cdot]$ is estimated by the empirical mean over the group $\mathbf{Y}$ of $G$ responses. 
The clipped surrogate term is given by
$\mathcal{J}_{\mathrm{clip},j,t}(\theta) = \min\big(\rho_{j,t}(\theta) A_j,
\mathrm{clip}(\rho_{j,t}(\theta), 1-\epsilon, 1+\epsilon)\, A_j\big)$, where
$\rho_{j,t}(\theta) = \pi_\theta(y_{j,t}|x,y_{j,<t})/\pi_{\theta_{\mathrm{old}}}(y_{j,t}|x,y_{j,<t})$ is the
importance ratio for the $t$-th token of the $j$-th sampled response $y_j$.
$\epsilon$ is the clipping parameter, and $A_j$ denotes the corresponding advantage, which is computed by normalizing the rewards within the sampled group:
\begin{equation}
A_j =
\frac{r(x,y_j)-\bar{r}}
{\sigma_r+\varepsilon},
\qquad
\bar{r}=\frac{1}{G}\sum_{k=1}^{G} r(x,y_k),
\end{equation}
where $\sigma_r$ denotes the standard deviation of rewards within the group and $\varepsilon$ is a small constant for numerical stability.


As in \Sref{sec:preliminaries}, the complete unlearning objective combines both terms:
\begin{equation}
\mathcal{L}_{\textnormal{unlearn-GRPO}}(\theta) = \mathcal{L}_{\textnormal{forget-GRPO}}(\theta; \mathcal{D}_{\mathrm{forget}}) + \lambda\,  \mathcal{L}_{\mathrm{retain}}(\theta; \mathcal{D}_{\mathrm{retain}}).
\label{eq:unlearn_obj}
\end{equation}

\paragraph{Reward Function.} Following the binary reward design of prior reinforcement unlearning work~\citep{zaradoukas2026reinforcementunlearninggrouprelative}, we adopt an LLM-as-a-judge~\citep{zheng2023judgingllmasajudgemtbenchchatbot} to assign two binary attributes to each sampled response $y_j$ to a forget query $x \in \mathcal{D}_{\mathrm{forget}}$: 
1) knowledge disclosure $k_j$, indicating whether the response reveals target algorithm knowledge; 
2) readability $u_j$, indicating whether the response remains fluent and understandable.
We define the reward as
\begin{equation}
r(x,y_j)=\mathds{1}\{(k_j,u_j)=(0,1)\}^{\footnotemark}.
\label{eq:reward}
\end{equation}
\footnotetext{$\mathds{1}\{\cdot\}$ is the indicator function that equals $1$ if the condition holds and $0$ otherwise.}%
That is, a response receives reward 1 only when it does not reveal target knowledge and remains readable; otherwise, the reward is 0.
The full judge prompts are provided in Appendix~\ref{appendix:judge_prompt_template}.


\begin{table*}[t]
\centering
\caption{Recovery Rate (RR, \%) and Coverage (\%) across algorithms,
models, and hint levels. Higher is better. Level 1 uses high-level hints;
level 2 uses step-by-step hints. Darker shading indicates
higher value. The order is based on reference difficulty ratings of algorithms (Appendix~\ref{app:cs_algos_rank}).
}
\label{tab:final_unlearned_success_rate}
\begingroup
\scriptsize\fontfamily{DejaVuSansMono-TLF}\selectfont
\setlength{\tabcolsep}{0pt}
\renewcommand{\arraystretch}{1.2}
\definecolor{rsrBlue}{RGB}{76,157,230}
\definecolor{rsrOrange}{RGB}{244,166,78}
\definecolor{rsrPurple}{RGB}{164,132,214}
\newcommand{\rsrarrow}[1]{%
  \makebox[8pt][c]{\tikz[x=1pt,y=1pt,baseline=-4.5pt]{%
    \draw[line width=0.55pt,line cap=round,
      -{Stealth[length=2.6pt,width=2.8pt]}] (0,0) -- #1;}}}
\newlength{\rsrlabelwidth}
\setlength{\rsrlabelwidth}{65pt}
\newlength{\rsrheadertextwidth}
\settowidth{\rsrheadertextwidth}{Algorithm}

\newcommand{\rsrheader}[2]{%
  \makebox[\rsrlabelwidth][l]{%
    \makebox[\rsrheadertextwidth][l]{#1}%
    \hspace{3pt}\rsrarrow{#2}}}
\newlength{\rsrwidth}
\setlength{\rsrwidth}{\dimexpr(\linewidth-\rsrlabelwidth-22pt)/9\relax}

\newcolumntype{R}{>{\centering\arraybackslash}p{\rsrwidth}} 

\newcommand{\rsrcell}[2]{%
  \begingroup
  \pgfmathsetmacro{\rsrtint}{8+0.32*#2}%
  \setlength{\fboxsep}{0pt}%
  \colorbox{#1!\rsrtint!white}{%
    \makebox[\rsrwidth][c]{\rule[-0.9ex]{0pt}{2.8ex}#2}}%
  \endgroup}

\newlength{\groupwidth}
\setlength{\groupwidth}{\dimexpr 3\rsrwidth + 2pt\relax}

\newcommand{\rsrB}[1]{\rsrcell{rsrBlue}{#1}}
\newcommand{\rsrO}[1]{\rsrcell{rsrOrange}{#1}}
\newcommand{\rsrP}[1]{\rsrcell{rsrPurple}{#1}}

\begin{tabular}{@{}p{\rsrlabelwidth}@{\hspace{4pt}}
R@{\hspace{1pt}}R@{\hspace{1pt}}R@{\hspace{6pt}}
R@{\hspace{1pt}}R@{\hspace{1pt}}R@{\hspace{6pt}}
R@{\hspace{1pt}}R@{\hspace{1pt}}R@{}}
\toprule
\rsrheader{RR (\%)}{(5,-5)}
& \multicolumn{3}{c@{\hspace{6pt}}}{\makebox[\groupwidth][c]{\shortstack{Qwen3-4B-Thinking}}}
& \multicolumn{3}{c@{\hspace{6pt}}}{\makebox[\groupwidth][c]{\shortstack{Qwen3-4B-Instruct}}}
& \multicolumn{3}{c@{}}{\makebox[\groupwidth][c]{\shortstack{Gemma-4-12B (Thinking)}}} \\
\cmidrule(r{6pt}){2-4}\cmidrule(r{6pt}){5-7}\cmidrule{8-10}
\rsrheader{Algorithm}{(0,-6)}
& \shortstack{no hint} & \shortstack{level-1} & \shortstack{level-2}
& \shortstack{no hint} & \shortstack{level-1} & \shortstack{level-2}
& \shortstack{no hint} & \shortstack{level-1} & \shortstack{level-2} \\
\midrule
Kadane
& \rsrB{96.1} & \rsrB{96.1} & \rsrB{100.0}
& \rsrP{82.8} & \rsrP{82.0} & \rsrP{75.0}
& \rsrO{98.4} & \rsrO{100.0} & \rsrO{99.2} \\
DAG
& \rsrB{70.3} & \rsrB{76.6} & \rsrB{100.0}
& \rsrP{28.1} & \rsrP{19.5} & \rsrP{25.0}
& \rsrO{93.8} & \rsrO{96.9} & \rsrO{100.0} \\
Gray
& \rsrB{50.8} & \rsrB{57.0} & \rsrB{100.0}
& \rsrP{0.0} & \rsrP{0.0} & \rsrP{75.0}
& \rsrO{40.6} & \rsrO{40.6} & \rsrO{99.2} \\
BFS
& \rsrB{35.9} & \rsrB{48.4} & \rsrB{59.4}
& \rsrP{0.0} & \rsrP{0.0} & \rsrP{75.0}
& \rsrO{31.2} & \rsrO{42.2} & \rsrO{100.0} \\
Euclidean
& \rsrB{37.5} & \rsrB{58.6} & \rsrB{86.7}
& \rsrP{1.6} & \rsrP{0.0} & \rsrP{87.5}
& \rsrO{37.5} & \rsrO{40.6} & \rsrO{67.2} \\
Moore Vote
& \rsrB{0.0} & \rsrB{0.0} & \rsrB{100.0}
& \rsrP{0.0} & \rsrP{5.5} & \rsrP{43.8}
& \rsrO{0.0} & \rsrO{3.1} & \rsrO{87.5} \\
Huffman
& \rsrB{93.8} & \rsrB{100.0} & \rsrB{100.0}
& \rsrP{0.0} & \rsrP{24.2} & \rsrP{81.2}
& \rsrO{99.2} & \rsrO{100.0} & \rsrO{100.0} \\
Union-Find
& \rsrB{96.9} & \rsrB{100.0} & \rsrB{98.4}
& \rsrP{10.2} & \rsrP{9.4} & \rsrP{37.5}
& \rsrO{81.2} & \rsrO{82.8} & \rsrO{87.5} \\
Floyd-Warshall
& \rsrB{42.2} & \rsrB{87.5} & \rsrB{85.2}
& \rsrP{0.8} & \rsrP{5.5} & \rsrP{93.8}
& \rsrO{96.9} & \rsrO{89.1} & \rsrO{90.6} \\
Dijkstra
& \rsrB{55.5} & \rsrB{70.3} & \rsrB{85.9}
& \rsrP{8.6} & \rsrP{35.9} & \rsrP{50.0}
& \rsrO{88.3} & \rsrO{95.3} & \rsrO{100.0} \\
Bellman-Ford
& \rsrB{22.7} & \rsrB{41.4} & \rsrB{96.1}
& \rsrP{0.0} & \rsrP{83.6} & \rsrP{87.5}
& \rsrO{34.4} & \rsrO{81.2} & \rsrO{96.1} \\
Prim
& \rsrB{0.0} & \rsrB{10.2} & \rsrB{100.0}
& \rsrP{0.0} & \rsrP{0.0} & \rsrP{87.5}
& \rsrO{3.1} & \rsrO{0.0} & \rsrO{93.8} \\
KMP
& \rsrB{0.0} & \rsrB{15.6} & \rsrB{56.2}
& \rsrP{0.0} & \rsrP{0.0} & \rsrP{31.2}
& \rsrO{0.0} & \rsrO{3.1} & \rsrO{85.9} \\
Hierholzer
& \rsrB{0.0} & \rsrB{0.0} & \rsrB{4.7}
& \rsrP{0.0} & \rsrP{0.0} & \rsrP{18.8}
& \rsrO{0.0} & \rsrO{0.0} & \rsrO{13.3} \\
Manacher
& \rsrB{0.0} & \rsrB{0.0} & \rsrB{31.2}
& \rsrP{0.0} & \rsrP{0.0} & \rsrP{25.0}
& \rsrO{0.0} & \rsrO{25.0} & \rsrO{15.6} \\
Stoer-Wagner
& \rsrB{0.0} & \rsrB{14.8} & \rsrB{22.7}
& \rsrP{0.0} & \rsrP{0.0} & \rsrP{0.0}
& \rsrO{0.0} & \rsrO{9.4} & \rsrO{6.2} \\
Pollard-Rho
& \rsrB{0.0} & \rsrB{0.0} & \rsrB{75.0}
& \rsrP{0.0} & \rsrP{0.8} & \rsrP{0.0}
& \rsrO{0.0} & \rsrO{0.0} & \rsrO{68.8} \\
Hungarian
& \rsrB{0.0} & \rsrB{0.0} & \rsrB{24.2}
& \rsrP{0.0} & \rsrP{0.0} & \rsrP{0.0}
& \rsrO{0.0} & \rsrO{0.0} & \rsrO{17.2} \\
BWT-Inverse
& \rsrB{0.0} & \rsrB{0.0} & \rsrB{100.0}
& \rsrP{0.0} & \rsrP{0.0} & \rsrP{37.5}
& \rsrO{0.0} & \rsrO{0.0} & \rsrO{48.4} \\
Strassen
& \rsrB{0.0} & \rsrB{0.0} & \rsrB{25.0}
& \rsrP{0.0} & \rsrP{0.0} & \rsrP{0.0}
& \rsrO{0.0} & \rsrO{0.0} & \rsrO{88.3} \\
\midrule
\textbf{Average RR (\%)}
& {\bfseries \rsrB{30.1}} & {\bfseries \rsrB{38.8}} & {\bfseries \rsrB{72.5}}
& {\bfseries \rsrP{6.6}}  & {\bfseries \rsrP{13.3}} & {\bfseries \rsrP{46.6}}
& {\bfseries \rsrO{35.2}} & {\bfseries \rsrO{40.5}} & {\bfseries \rsrO{73.2}} \\
\textbf{Coverage (\%)}
& {\bfseries \rsrB{50.0}} & {\bfseries \rsrB{65.0}} & {\bfseries \rsrB{100.0}}
& {\bfseries \rsrP{30.0}} & {\bfseries \rsrP{45.0}} & {\bfseries \rsrP{80.0}}
& {\bfseries \rsrO{55.0}} & {\bfseries \rsrO{70.0}} & {\bfseries \rsrO{100.0}} \\
\bottomrule
\end{tabular}

\endgroup
\end{table*}

\paragraph{Cold Start.}
\label{sec:cold_start}
In our early attempts, at the beginning of on-policy unlearning, initial model responses to queries in $\mathcal{D}_{\mathrm{forget}}$ typically receive zero reward, leaving little effective optimization signal for policy updates.
To address this, we apply a cold start stage~\citep{zhang2025rulereinforcementunlearningachieves}: we construct an initialization set $\mathcal{D}_{\mathrm{init}}=\{(x_i,y_i)\}$ of target-related queries paired with refusal-style responses in the model's original style.
We use $\mathcal{D}_{\mathrm{init}}$ to guide the model toward refusal-style behavior on forget queries via supervised fine-tuning. Details are provided in Appendix~\ref{app:data_details}.

\paragraph{Algorithm-name suppression.}
We also try a weaker form of ``forgetting'' that suppresses algorithm-name generation by masking selected name-related tokens during decoding, without updating model parameters.
Mask construction and decoding details are provided in Appendix~\ref{app:name_suppression}.

\begin{figure*}[t]
\centering
\includegraphics[width=1.0\textwidth]{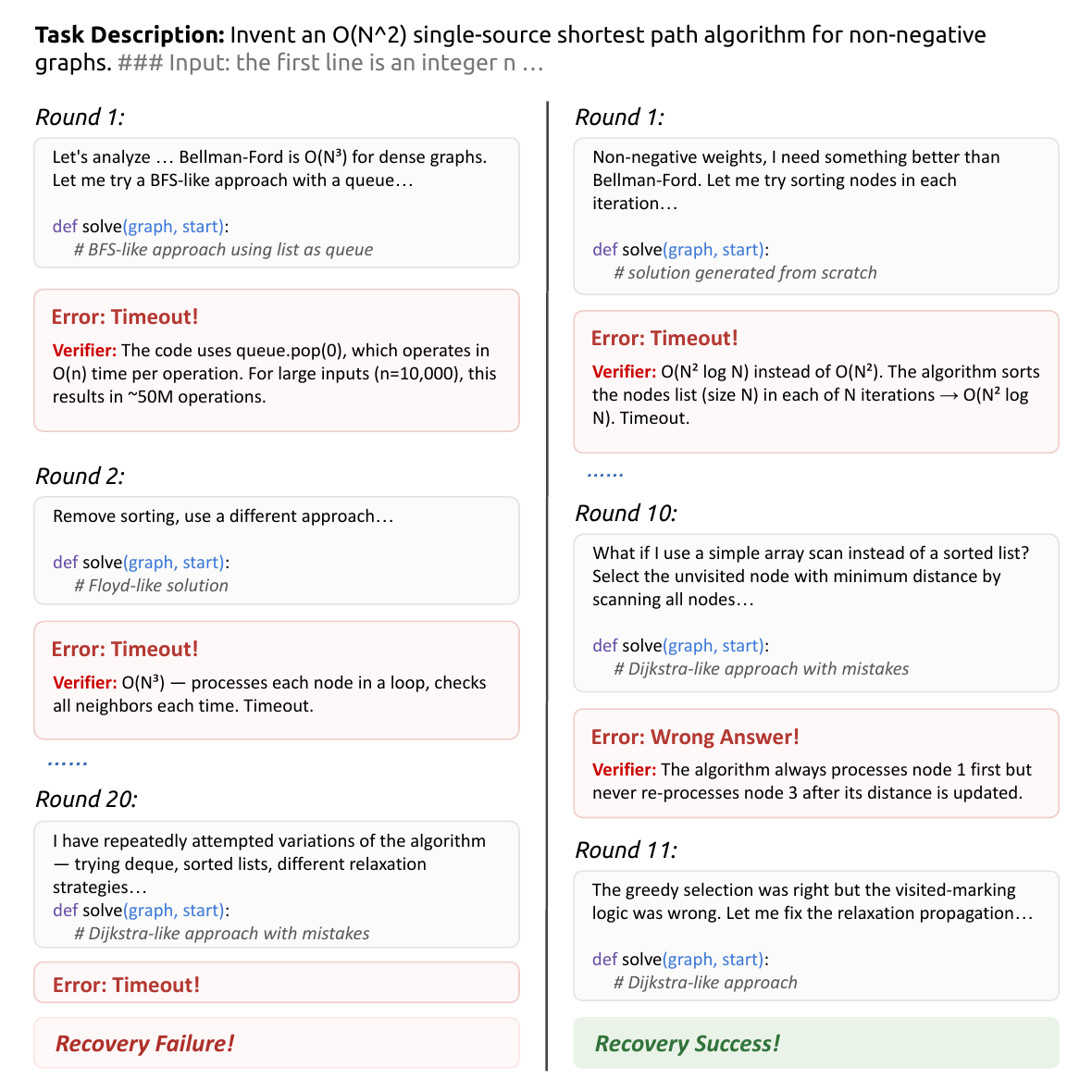}
\caption{ 
Two recovery trajectories for Dijkstra's algorithm (no hint) on Qwen3-4B-Thinking. \textbf{Left:} the model remains stuck in suboptimal strategies and eventually fails. \textbf{Right:} the model comes up with a Dijkstra-like solution by round 11, corrected through verifier feedback.
}
\label{fig:main3}
\vspace{-15pt}
\end{figure*}

\subsection{The Recovery Phase}
\label{sec:reinvention_phase}

\begin{figure*}[t]
\centering
\includegraphics[width=\textwidth]{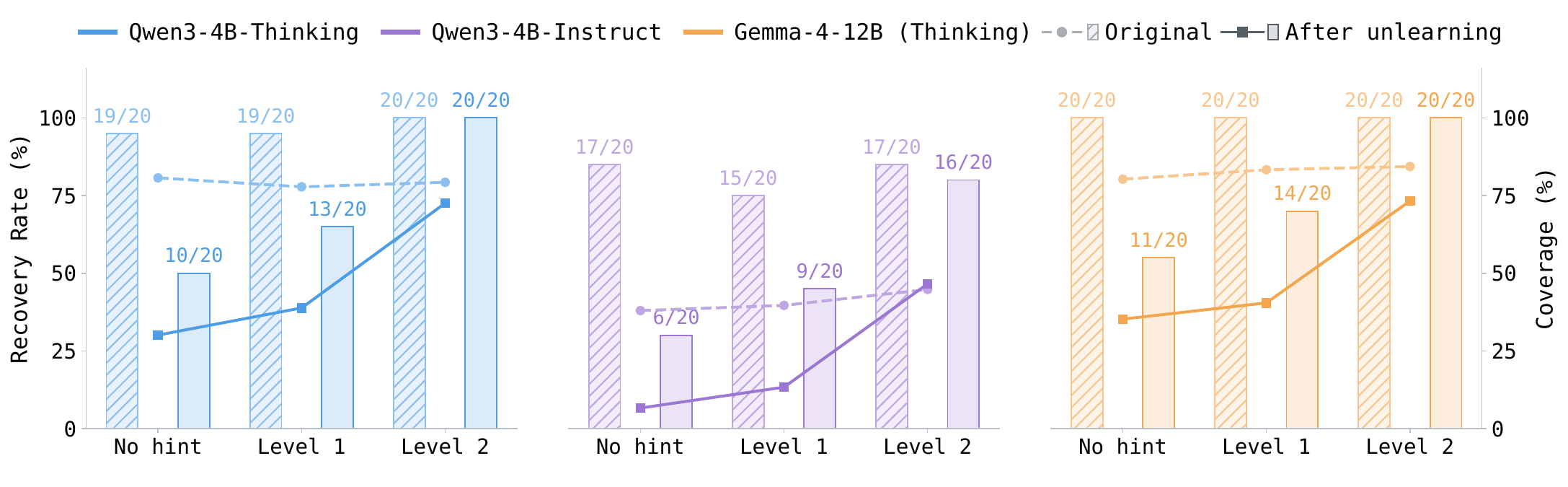}
\vspace{-20pt}
\caption{
Recovery rate (lines) and algorithm coverage (bars)
before and after unlearning across hint levels.
More informative hints substantially narrow the recovery gap between
unlearned and original models.
With level-2 hints, unlearned Qwen3-4B-Instruct even achieves a better recovery rate than its original counterparts.
}
\vspace{-15pt}
\label{fig:recovery_overview}
\end{figure*}

\paragraph{Environment.}
\label{sec:env}
After unlearning, we evaluate whether the model can recover the target algorithm by interacting with a Python interpreter. For a target algorithm, we define an application task with a prompt, test suite and runtime limit. The prompt presents a programming problem with efficiency requirements tailored to the target algorithm or its variants. The unlearned model interacts with the Python interpreter by executing candidate code and eventually submitting a final solution for evaluation. We discuss evaluation validity in Appendix~\ref{app:eval_validity}.

A submission is considered successful only if it passes the test suite within the runtime limit, which is calibrated according to the time complexity of the target algorithm. Calibration details are provided in Appendix~\ref{app:runtime_calibration}.

In preliminary experiments, we identify a failure mode during algorithm recovery after unlearning: after successive failed attempts, models usually produce shorter reasoning traces and abandon unresolved tasks.
Motivated by recent AI research systems that employ generative verifiers at test time~\citep{feng2026autonomousmathematicsresearch,feng2026aletheiatacklesfirstproofautonomously}, we incorporate a generative verifier~\citep{zhang2025generativeverifiersrewardmodeling} for failed submissions.
By default, this verifier is instantiated by the unlearned model itself and returns diagnostic feedback to guide the model's revision.
The verifier prompt template is provided in Appendix~\ref{appendix:verifier_prompt_template}.
We examine the role of the verifier through ablation experiments in \Sref{sec:ablation_verify}.

\paragraph{Hint Levels.}
To study how external hints affect recovery, we evaluate models under three prompt levels that vary in the amount of guidance provided: no hint beyond the task description, level 1 (a high-level conceptual hint about the algorithmic approach), and level 2 (a detailed step-by-step explanation in natural language).
Hint examples are provided in Appendix~\ref{appendix:prompt_level_templates}.

\vspace{-5pt}

\paragraph{Remarks.}
One might worry that unlearning impairs algorithm-specific capabilities (e.g., by inducing persistent refusal-type responses on tasks involving the target algorithm).
To reduce this risk, we restrict our unlearning data to concept-related queries, excluding the application tasks used during recovery.
Meanwhile, in subsequent experiments, algorithmic hints substantially narrow the performance gap between unlearned and original models (\Sref{sec:result_hint}).
This suggests that the unlearned models retain substantial coding and implementation capabilities in algorithm-specific settings.

\vspace{-5pt}

\section{Experimental Settings}
\vspace{-5pt}

\label{sec:exp_setup}

\paragraph{Models.}
We select 3 strong open-weight models with 4B to 12B parameters as backbones for our unlearning and recovery experiments: Qwen3-4B-Thinking-2507, Qwen3-4B-Instruct-2507~\citep{qwen3technicalreport}, and Gemma-4-12B~\citep{gemmateam2026gemma4}.
Larger frontier open-source models, Qwen3.8-27B~\citep{qwen38} and Gemma-4-31B~\citep{gemmateam2026gemma4}, are also included in the algorithm-name suppression experiments.
We use DeepSeek-V4-Flash~\citep{deepseekai2026deepseekv4} for all LLM-as-a-judge components, with a sampling temperature of 1.0.
Training details for each model are provided in Appendix~\ref{app:unlearning}.

\paragraph{Evaluation.}
Following prior work on probing-based unlearning evaluation~\citep{sun2025unlearningvsobfuscationtruly}, we construct multiple-choice and open-ended questions for each target algorithm. 
We adapt the judge-based knowledge reproduction assessment of~\citet{tan2026wisdomknowingsayhallucinationfree}.
For each probe, the judge model examines both the reasoning process and the final response, assigning a label $y \in \{0, 1\}$ where $y=1$ indicates that the model reveals no knowledge of the target algorithm.
The Forgetting Rate (FR) is then defined as $\text{FR} = \frac{1}{N}\sum_{i=1}^{N} y_i$, where $N$ is the total number of probes.

We evaluate retained model utility on LiveCodeBench~\citep{jain2024livecodebenchholisticcontaminationfree} and AIME25~\citep{aime25}.
Following~\citet{qwen3_huggingface}, we use LiveCodeBench v6 (date: 25.02--25.05).
For recovery, we report Recovery Rate (RR), defined as the fraction of successful attempts for a target algorithm, and Coverage, defined as the fraction of target algorithms
with at least one successful attempt under the evaluation budget.
For each algorithm, we construct 8 problem variants and run 128 trials at each prompt level across these variants.
More evaluation settings are provided in Appendix~\ref{app:eval}.

\paragraph{Datasets.}
For data construction, the target-specific queries in $\mathcal{D}_{\mathrm{init}}$, $\mathcal{D}_{\mathrm{forget}}$, and the evaluation probe sets are generated by DeepSeek-V4-Flash~\citep{deepseekai2026deepseekv4}.
The queries in $\mathcal{D}_{\mathrm{retain}}$ are sampled from the \texttt{Code} and \texttt{Math} categories of Nemotron-Post-Training-Dataset-v1~\citep{NemotronPostTrainingDatasetV1, bercovich2025llamanemotronefficientreasoningmodels}.
Appendix~\ref{app:data_details} describes the dataset construction process and provides dataset statistics, prompt templates, and examples.

\vspace{-5pt}

\section{Experiments}

\subsection{Main Results}
\vspace{-5pt}
\label{sec:recovery_after_unlearning}

In this section, we compare the unlearned models (Table~\ref{tab:final_unlearned_success_rate}) with their original counterparts (Table~\ref{tab:base}), focusing on Recovery Rates and Coverages (introduced in \Sref{sec:exp_setup}).

\paragraph{Nearly Half of the Target Algorithms Remain Unrecovered without Hints.}
As shown in Table~\ref{tab:final_unlearned_success_rate} and Table~\ref{tab:base} in Appendix~\ref{app:base}, we observe a substantial gap between the original models' task performance and their recovery rate after unlearning.
Before unlearning, the two thinking models could successfully solve almost all algorithms in our setting.
However, although unlearning does not guarantee complete removal of target-algorithm knowledge, 9 of the 20 foundational algorithms have no successful recovery across the three evaluated models without hints.

Recovery also exhibits a clear pattern with respect to the difficulty ratings of the algorithms. 
We use the Luogu ratings~\citep{luogu_difficulty} reported in Appendix~\ref{app:cs_algos_rank} as a coarse proxy for algorithm difficulty.
Among the 12 algorithms whose template problems are rated at levels 2--3, 11 are recovered by at least one model without hints.
In contrast, none of the 6 algorithms with template problems rated at levels 4--7 are recovered under the same no-hint setting. 
This gap suggests that recovery becomes harder as algorithms grow more complex, suggesting models' limited ability to achieve innovation on difficult foundational algorithms.
To better understand model behavior during recovery, we provide a representative success and failure trajectory for Dijkstra's algorithm in Figure~\ref{fig:main3}, with more examples in Appendix~\ref{app:traj}.

\paragraph{High Forgetting Rates Coexist with Substantial Retained Utility.}
As shown in Table~\ref{tab:final_unlearned_metrics}, the average forgetting rates over the reported targets are 96.0\%, 94.0\%, and 90.0\% for Qwen3-4B-Thinking, Qwen3-4B-Instruct, and Gemma-4-12B, respectively, indicating that most probe responses are judged not to reveal target-algorithm knowledge.
Meanwhile, the unlearned models retain substantial performance on LCB and AIME25.
These results show that recovery failures coexist with substantial retained performance on broader reasoning tasks.

\paragraph{Algorithmic Hints Narrow the Recovery Gap.}
\label{sec:result_hint}

\begin{figure*}[t]
  \centering
  \includegraphics[width=\textwidth]{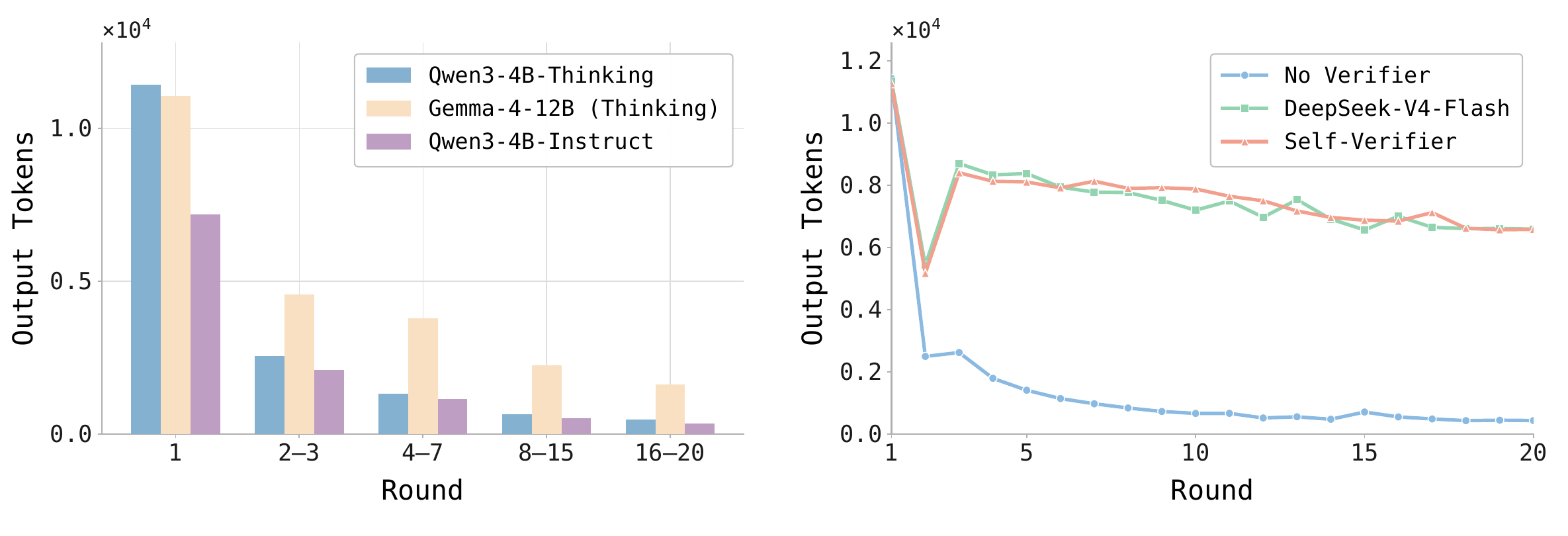}
\vspace{-20pt}
  \caption{
Round-wise output tokens during recovery.
\textbf{Left}: Across three models without verifier feedback, outputs become shorter in later rounds.
\textbf{Right}: Three verifier settings on Qwen3-4B-Thinking. 
Without a verifier, the model either succeeds early or quickly abandons the task.
With a verifier, the model sustains meaningful exploration over more rounds, enabling success even when the correct approach is not found initially.
  }
\label{fig:length_round}
\vspace{-15pt}
\end{figure*}

As shown in Figure~\ref{fig:recovery_overview}, for each of the three backbone models, increasingly detailed hints improve average recovery rates and algorithm coverage after unlearning, narrowing the gap relative to the original model.
Compared with the no-hint setting, level-1 hints enable recovery of additional algorithms, while level-2 hints yield larger overall gains.
Notably, at level 2, the unlearned variants of Qwen3-4B-Instruct achieve a higher average recovery rate than the original model.
Across hint levels, we also observe successful recovery on some algorithm--hint combinations for which the corresponding original models produce nearly no successful solutions.
For example, under level-2 hints, unlearned Qwen3-4B-Instruct variants achieve recovery rates of 37.5\% and 18.8\% on BWT-Inverse and Hierholzer, respectively, compared with 0.8\% and 0.0\% for the original model.
These results show that the unlearned models can effectively use explicit algorithmic guidance, in some cases even achieving successful ``recovery'' beyond the original models' observed coverage.

\subsection{Recovery under Algorithm-Name Suppression}
\label{sec:name_suppression}
Beyond parameter-based unlearning, we evaluate algorithm-name suppression on the four thinking models across all 20 algorithms without hints.


\paragraph{Algorithm-name suppression substantially reduces recovery.}
\begin{wrapfigure}{r}{0.48\textwidth}
\centering
\includegraphics[width=\linewidth]{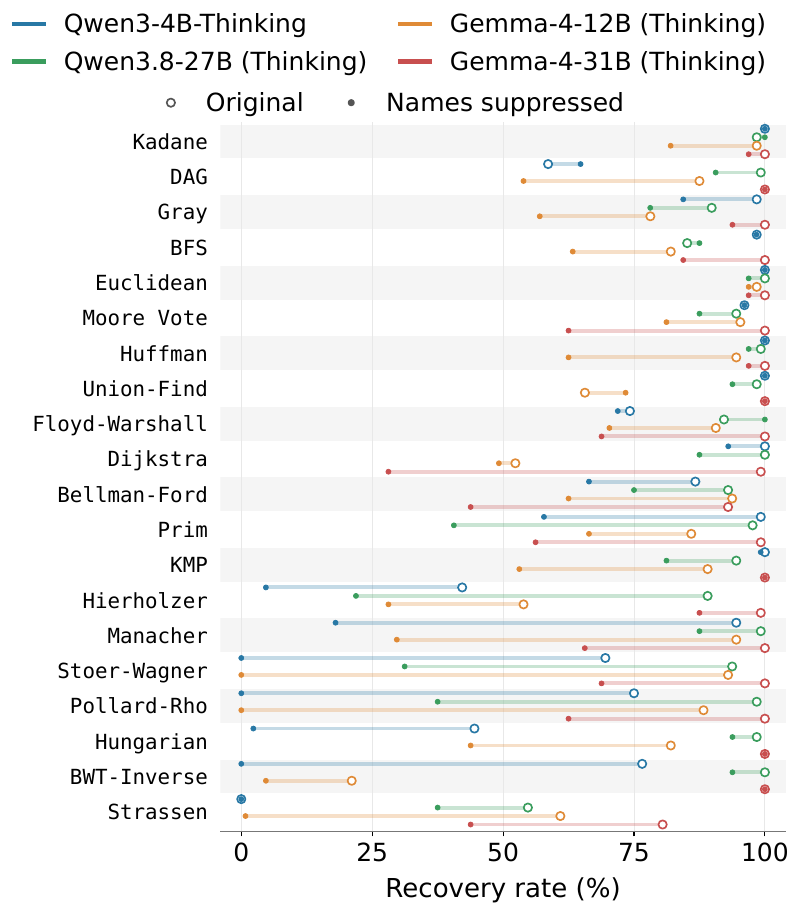}
\caption{
Effect of algorithm-name suppression on four thinking models without hints.
Each line links an algorithm's original recovery rate to its rate after name suppression. Line length indicates the absolute change.
}
\label{fig:name_suppression}
\vspace{-\intextsep}
\vspace{-\intextsep}
\end{wrapfigure}

As shown in Figure~\ref{fig:name_suppression}, algorithm-name suppression reduces mean recovery rates by an average of 23.2 percentage points, with a mean per-model relative decline of 26.9\%.
Mean algorithm coverage decreases from 98.8\% to 92.5\%.
On the other hand, performance on broader reasoning benchmarks remains strong (Tables~\ref{tab:suppression_utility} and~\ref{tab:suppression_utility_additional}, Appendix~\ref{app:suppression_utility}).

\paragraph{Algorithm names serve as retrieval cues.}
The recovery rate degradation suggests the important role of name-associated retrieval in such tasks.
Without generating the algorithm's name, the model must express a solution through alternative descriptions or directly in code.
Notably, we observe that large declines occur for algorithms such as Manacher, Stoer--Wagner, and Pollard--Rho, while simpler algorithms such as Euclidean remain stable.
This might be due to their proportion in pretraining data or their difficulty, making such alternative routes less accessible for these harder algorithms.
These findings suggest the model's overreliance on name-associated memory when solving foundational algorithmic problems.
More case studies and behaviour analyses are provided in Appendix~\ref{app:name_suppression}.



\vspace{-3pt}
\subsection{Ablation of the Generative Verifier}
\vspace{-3pt}
\label{sec:ablation_verify}
\begin{wraptable}{r}{0.46\textwidth}
\vspace{-\intextsep}
\centering

\caption{
Verifier ablation on Qwen3-4B-Thinking in the no-hint setting.
Recovery Rate (RR, \%) and Coverage (\%) are averaged across 20 algorithms.
}
\label{tab:rsr_verify}

\begingroup
\scriptsize\fontfamily{DejaVuSansMono-TLF}\selectfont
\renewcommand{\arraystretch}{1.08}
\setlength{\tabcolsep}{6pt}

\begin{tabular}{@{}lcc@{}}
\toprule
Verifier & RR (\%) & Coverage (\%) \\
\midrule
No Verifier & 20.5 & 50.0 \\
DeepSeek-V4-Flash & 36.6 & 50.0 \\
Self-Verifier & {30.1} & 50.0 \\
\bottomrule
\end{tabular}

\endgroup
\vspace{-\intextsep}
\end{wraptable}
To isolate the contribution of the generative verifier, we remove verifier feedback and analyze recovery trajectories across three models. 
Without feedback, model outputs shorten over rounds (Figure~\ref{fig:length_round}), reflecting reduced exploration---later rounds contain fewer candidate ideas, and in some cases the model abandons problem-solving entirely or attributes failures to the testing environment rather than its own solution. 
This suggests limitations in the models' ability to reflect on failures and sustain reasoning across multiple interaction rounds.
We refer to this phenomenon as \emph{thought collapse}; additional examples are provided in Appendix~\ref{app:thought_collapse}.

We then test whether verifier feedback can mitigate ``thought collapse'' on Qwen3-4B-Thinking.
We compare three settings: no verifier, the default self-verifier, and a strong verifier implemented with DeepSeek-V4-Flash.
Compared with the no-feedback setting, verifiers help maintain longer output lengths across rounds (Figure~\ref{fig:length_round}).
These gains are also reflected in a higher recovery rate (Table~\ref{tab:rsr_verify}), suggesting that natural-language failure diagnosis helps sustain exploration and improve recovery outcomes.
However, verifiers improve recovery frequency without increasing the number of recovered algorithms.
We further investigate this pattern by analyzing how coverage varies with the number of trials and interaction turns in Appendix~\ref{app:recovery-scaling}.

\vspace{-1.5pt}



\begin{table*}[t]
\centering
\caption{
Unlearning results across target algorithms and models.
Forgetting Rate (FR, \%) measures unlearning effectiveness,
while LCB and AIME25 (detailed in \Sref{sec:exp_setup})
measure model utility.
\textit{Origin} denotes the original model before unlearning.
}
\label{tab:final_unlearned_metrics}
\vspace{-5pt}

\begingroup
\scriptsize
\fontfamily{DejaVuSansMono-TLF}\selectfont
\setlength{\tabcolsep}{0pt}
\renewcommand{\arraystretch}{1.2}

\setlength{\rsrwidth}{\dimexpr(\linewidth-96pt)/9\relax}

\setlength{\groupwidth}{\dimexpr 3\rsrwidth\relax}

\newcolumntype{R}{>{\centering\arraybackslash}p{\rsrwidth}}

\begin{tabular}{@{}p{96pt} R R R R R R R R R @{}}
\toprule
\multirow{2}{*}{Algorithm}
& \multicolumn{3}{c}{\makebox[\groupwidth][c]{\shortstack{Qwen3-4B-Thinking}}}
& \multicolumn{3}{c}{\makebox[\groupwidth][c]{\shortstack{Qwen3-4B-Instruct}}}
& \multicolumn{3}{c}{\makebox[\groupwidth][c]{\shortstack{Gemma-4-12B (Thinking)}}} \\
\cmidrule(lr){2-4}\cmidrule(lr){5-7}\cmidrule{8-10}
& FR (\%) & LCB & AIME25
& FR (\%) & LCB & AIME25
& FR (\%) & LCB & AIME25 \\
\midrule
\rowcolor{gray!15}
\textit{Origin}
& 0.0 & 55.2 & 81.3
& 0.0 & 35.1 & 47.4
& 0.0 & 71.8 & 82.2 \\

Kadane
& 100.0 & 50.4 & 78.9
& 96.6 & 32.1 & 47.8
& 94.9 & 63.4 & 68.9 \\

DAG
& 86.7 & 48.1 & 80.0
& 96.7 & 38.9 & 47.8
& 91.7 & 62.6 & 70.0 \\

Gray
& 97.1 & 50.4 & 74.4
& 91.4 & 38.9 & 47.8
& 88.6 & 66.4 & 73.3 \\

BFS
& 91.7 & 51.1 & 71.1
& 96.7 & 41.2 & 45.6
& 85.0 & 72.5 & 71.1 \\

Euclidean
& 100.0 & 50.4 & 74.4
& 95.0 & 43.5 & 40.0
& 80.0 & 67.9 & 72.2 \\

Moore Vote
& 97.4 & 48.9 & 78.9
& 97.4 & 38.2 & 46.7
& 92.3 & 66.4 & 71.1 \\

Huffman
& 98.3 & 49.6 & 78.9
& 95.0 & 35.9 & 44.4
& 88.3 & 71.0 & 72.2 \\

Union-Find
& 96.7 & 48.1 & 77.8
& 98.3 & 31.3 & 43.3
& 90.0 & 68.7 & 72.2 \\

Floyd-Warshall
& 100.0 & 54.2 & 76.7
& 87.5 & 35.9 & 47.8
& 85.0 & 65.6 & 73.3 \\

Dijkstra
& 72.1 & 49.6 & 76.7
& 82.0 & 33.6 & 38.9
& 73.8 & 68.7 & 67.8 \\

Bellman-Ford
& 92.5 & 54.2 & 76.7
& 90.0 & 30.5 & 44.4
& 87.5 & 72.5 & 68.9 \\

Prim
& 100.0 & 48.9 & 78.9
& 85.0 & 35.9 & 42.2
& 87.5 & 74.8 & 76.7 \\

KMP
& 95.0 & 48.1 & 75.6
& 90.0 & 32.8 & 44.4
& 98.5 & 67.2 & 66.7 \\

Hierholzer
& 100.0 & 46.6 & 72.2
& 96.7 & 32.1 & 46.7
& 96.4 & 61.1 & 62.2 \\

Manacher
& 100.0 & 49.6 & 76.7
& 92.0 & 40.5 & 44.4
& 90.0 & 64.9 & 63.3 \\

Stoer-Wagner
& 100.0 & 47.3 & 76.7
& 98.3 & 35.9 & 42.2
& 90.0 & 66.4 & 73.3 \\

Pollard-Rho
& 100.0 & 50.4 & 77.8
& 98.3 & 38.9 & 43.3
& 95.0 & 64.1 & 66.7 \\

Hungarian
& 93.3 & 48.1 & 80.0
& 95.0 & 38.2 & 45.6
& 95.0 & 51.9 & 53.3 \\

BWT-Inverse
& 100.0 & 48.9 & 78.9
& 100.0 & 36.6 & 41.1
& 97.0 & 55.0 & 45.6 \\

Strassen
& 100.0 & 48.9 & 81.1
& 98.0 & 35.1 & 43.3
& 94.0 & 64.1 & 71.1 \\

\midrule
\textbf{Average}
& \textbf{96.0} & \textbf{49.6} & \textbf{77.1}
& \textbf{94.0} & \textbf{36.3} & \textbf{44.4}
& \textbf{90.0} & \textbf{65.8} & \textbf{68.0} \\
\bottomrule
\end{tabular}

\endgroup
\end{table*}




\section{Related Work}
\label{sec:related_work}
\vspace{-3pt}
\vspace{-3pt}

We discuss the most relevant studies here and provide a more comprehensive review in Appendix~\ref{app:more_related}.
Recent LLM-based systems have shown promise in scientific research and algorithmic discovery~\citep{FunSearch2023,novikov2025alphaevolvecodingagentscientific,gottweis2025aicoscientist}.
Our work examines whether models can recover foundational algorithms when direct recall of the target knowledge is suppressed.
LLM unlearning provides a practical intervention for this purpose, with existing methods based on gradient ascent, preference optimization, and on-policy reinforcement learning~\citep{jang-etal-2023-knowledge,zhang2024negativepreferenceoptimizationcatastrophic,zhang2025rulereinforcementunlearningachieves, zaradoukas2026reinforcementunlearninggrouprelative}.
Most closely related, \citet{yang2025unlearningablationfalsifiablebenchmark} proposes using unlearning as an ablation probe to investigate whether LLMs can re-derive scientific results after targeted knowledge removal.
We develop this conceptual proposal into a concrete experimental framework, examining recovery under different hint levels and verifier settings, alongside a complementary algorithm-name suppression intervention.

\vspace{-3pt}
\vspace{-3pt}
\section{Conclusion and Discussion}
\vspace{-3pt}
\vspace{-3pt}

In this work, we study whether LLMs can recover foundational algorithms when direct recall is suppressed.
Our experiments reveal two main findings: First, although unlearning does not guarantee complete knowledge removal, nearly half of the algorithms remain unrecovered without hints.
Second, suppressing algorithm-name-related tokens alone reduces recovery without modifying model parameters.
These results indicate LLMs' overreliance on existing knowledge and familiar retrieval cues.

It is also worth noting that we observe ``thought collapse'' and introduce a verifier to mitigate it.
Though verifier feedback sustains reasoning and improves recovery rates, continued revision does not expand algorithm coverage within our evaluation budget.
This suggests limitations in self-reflection and its ability to broaden algorithmic exploration.
\textbf{By treating the unlearned model's recovery rate as an approximate upper bound, our results suggest that current LLMs still have limited ability to make foundational algorithm innovation.}
We hope our findings could provide a new perspective on the gap between knowledge-based algorithm solving and algorithmic discovery, and offer insights into both the potential and current limits of LLMs' innovative reasoning.

\vspace{-3pt}
\vspace{-3pt}
\section{Limitations and Future Work}
\vspace{-3pt}
\vspace{-3pt}

We conclude with several limitations and directions for future work. 
Our experiments cover a limited set of settings and algorithms; it remains unclear whether our findings generalize to other scientific domains or larger models.
Meanwhile, unlearning does not guarantee equivalence to a model retrained without the target knowledge.
Future work may use retraining with controlled exclusion of target knowledge or alternative interventions, and extend the evaluation to larger models and broader discovery tasks.

\section*{AI use statement}
\vspace{-5pt}
We used generative AI tools for language polishing, literature searches, and LaTeX formatting.
These tools were also used to obtain critical feedback on the research methodology and the interpretation of results.
We also used DeepSeek-V4-Flash to generate target-specific queries for the unlearning datasets and evaluation probe sets, as detailed in Appendix~\ref{app:data_details}.
The authors take responsibility for the final content of this paper.

\vspace{-5pt}




\vspace{-3pt}
\section*{Acknowledgement}
\vspace{-3pt}
We sincerely thank Jingyu Zhang for guidance on paper writing and valuable feedback.

\bibliography{colm2026_conference}
\bibliographystyle{iclr2027_conference}


\appendix
\section*{Appendix}

\section{Additional Results}
\label{app:additional_analysis}


\begin{table*}[t]
\centering
\caption{Non-unlearned Recovery Rate (RR, \%) and Coverage (\%) across
target algorithms, models, and hint levels. Higher is better.
Level 1 uses high-level hints; level 2 uses step-by-step hints.
Each reported cell is based on 128 samples.
Darker shading indicates higher value.}
\label{tab:base}

\begingroup
\scriptsize\fontfamily{DejaVuSansMono-TLF}\selectfont
\setlength{\tabcolsep}{0pt}
\renewcommand{\arraystretch}{1.2}

\definecolor{baseBlue}{RGB}{76,157,230}
\definecolor{baseOrange}{RGB}{244,166,78}
\definecolor{basePurple}{RGB}{164,132,214}

\newcommand{\basearrow}[1]{%
  \makebox[8pt][c]{\tikz[x=1pt,y=1pt,baseline=-4.5pt]{%
    \draw[line width=0.55pt,line cap=round,
      -{Stealth[length=2.6pt,width=2.8pt]}] (0,0) -- #1;}}}

\newlength{\baselabelwidth}
\setlength{\baselabelwidth}{65pt}

\newlength{\baseheadertextwidth}
\settowidth{\baseheadertextwidth}{Algorithm}

\newcommand{\baseheader}[2]{%
  \makebox[\baselabelwidth][l]{%
    \makebox[\baseheadertextwidth][l]{#1}%
    \hspace{3pt}\basearrow{#2}}}

\newlength{\basewidth}
\setlength{\basewidth}{\dimexpr(\linewidth-90pt)/9\relax}
\newcolumntype{B}{>{\centering\arraybackslash}p{\basewidth}}

\newcommand{\basecell}[2]{%
  \begingroup
  \pgfmathsetmacro{\basetint}{8+0.32*#2}%
  \setlength{\fboxsep}{0pt}%
  \colorbox{#1!\basetint!white}{%
    \makebox[\basewidth][c]{\rule[-0.9ex]{0pt}{2.8ex}#2}}%
  \endgroup}

\newlength{\basegroupwidth}
\setlength{\basegroupwidth}{\dimexpr 3\basewidth + 2pt\relax}

\newcommand{\baseB}[1]{\basecell{baseBlue}{#1}}
\newcommand{\baseP}[1]{\basecell{basePurple}{#1}}
\newcommand{\baseO}[1]{\basecell{baseOrange}{#1}}

\begin{tabular}{@{}p{\baselabelwidth}@{\hspace{4pt}}
B@{\hspace{1pt}}B@{\hspace{1pt}}B@{\hspace{6pt}}
B@{\hspace{1pt}}B@{\hspace{1pt}}B@{\hspace{6pt}}
B@{\hspace{1pt}}B@{\hspace{1pt}}B@{}}
\toprule
\baseheader{RR (\%)}{(5,-5)}
& \multicolumn{3}{c@{\hspace{6pt}}}{\makebox[\basegroupwidth][c]{\shortstack{Qwen3-4B-Thinking}}}
& \multicolumn{3}{c@{\hspace{6pt}}}{\makebox[\basegroupwidth][c]{\shortstack{Qwen3-4B-Instruct}}}
& \multicolumn{3}{c@{}}{\makebox[\basegroupwidth][c]{\shortstack{Gemma-4-12B (Thinking)}}} \\
\cmidrule(r{6pt}){2-4}\cmidrule(r{6pt}){5-7}\cmidrule{8-10}
\baseheader{Algorithm}{(0,-6)}
& \shortstack{no hint} & \shortstack{level-1} & \shortstack{level-2}
& \shortstack{no hint} & \shortstack{level-1} & \shortstack{level-2}
& \shortstack{no hint} & \shortstack{level-1} & \shortstack{level-2} \\
\midrule
Kadane
& \baseB{100.0} & \baseB{100.0} & \baseB{100.0}
& \baseP{99.2} & \baseP{91.4} & \baseP{96.1}
& \baseO{98.4} & \baseO{99.2} & \baseO{100.0} \\
DAG
& \baseB{58.6} & \baseB{67.2} & \baseB{33.6}
& \baseP{21.9} & \baseP{14.8} & \baseP{13.3}
& \baseO{87.5} & \baseO{91.4} & \baseO{89.8} \\
Gray
& \baseB{98.4} & \baseB{92.2} & \baseB{100.0}
& \baseP{53.9} & \baseP{28.1} & \baseP{75.0}
& \baseO{78.1} & \baseO{84.4} & \baseO{97.7} \\
BFS
& \baseB{98.4} & \baseB{97.7} & \baseB{100.0}
& \baseP{9.4} & \baseP{7.0} & \baseP{1.6}
& \baseO{82.0} & \baseO{85.9} & \baseO{86.7} \\
Euclidean
& \baseB{100.0} & \baseB{100.0} & \baseB{100.0}
& \baseP{98.4} & \baseP{97.7} & \baseP{91.4}
& \baseO{98.4} & \baseO{100.0} & \baseO{99.2} \\
Moore Vote
& \baseB{96.1} & \baseB{95.3} & \baseB{100.0}
& \baseP{63.3} & \baseP{33.6} & \baseP{72.7}
& \baseO{95.3} & \baseO{91.4} & \baseO{85.9} \\
Huffman
& \baseB{100.0} & \baseB{100.0} & \baseB{100.0}
& \baseP{83.6} & \baseP{65.6} & \baseP{60.2}
& \baseO{94.5} & \baseO{94.5} & \baseO{93.0} \\
Union-Find
& \baseB{100.0} & \baseB{100.0} & \baseB{99.2}
& \baseP{21.1} & \baseP{24.2} & \baseP{28.9}
& \baseO{65.6} & \baseO{53.1} & \baseO{57.8} \\
Floyd-Warshall
& \baseB{74.2} & \baseB{69.5} & \baseB{68.8}
& \baseP{51.6} & \baseP{57.8} & \baseP{48.4}
& \baseO{90.6} & \baseO{91.4} & \baseO{91.4} \\
Dijkstra
& \baseB{100.0} & \baseB{100.0} & \baseB{100.0}
& \baseP{2.3} & \baseP{89.8} & \baseP{88.3}
& \baseO{52.3} & \baseO{72.7} & \baseO{84.4} \\
Bellman-Ford
& \baseB{86.7} & \baseB{71.1} & \baseB{59.4}
& \baseP{100.0} & \baseP{96.9} & \baseP{95.3}
& \baseO{93.8} & \baseO{89.1} & \baseO{72.7} \\
Prim
& \baseB{99.2} & \baseB{98.4} & \baseB{100.0}
& \baseP{15.6} & \baseP{21.9} & \baseP{76.6}
& \baseO{85.9} & \baseO{87.5} & \baseO{86.7} \\
KMP
& \baseB{100.0} & \baseB{100.0} & \baseB{100.0}
& \baseP{93.8} & \baseP{96.9} & \baseP{91.4}
& \baseO{89.1} & \baseO{93.0} & \baseO{93.8} \\
Hierholzer
& \baseB{42.2} & \baseB{35.2} & \baseB{50.0}
& \baseP{0.0} & \baseP{0.0} & \baseP{0.0}
& \baseO{53.9} & \baseO{56.2} & \baseO{71.1} \\
Manacher
& \baseB{94.5} & \baseB{98.4} & \baseB{99.2}
& \baseP{40.6} & \baseP{59.4} & \baseP{49.2}
& \baseO{94.5} & \baseO{95.3} & \baseO{93.8} \\
Stoer-Wagner
& \baseB{69.5} & \baseB{75.8} & \baseB{75.8}
& \baseP{1.6} & \baseP{0.0} & \baseP{0.8}
& \baseO{93.0} & \baseO{96.1} & \baseO{93.0} \\
Pollard-Rho
& \baseB{75.0} & \baseB{54.7} & \baseB{63.3}
& \baseP{1.6} & \baseP{7.8} & \baseP{5.5}
& \baseO{88.3} & \baseO{86.7} & \baseO{90.6} \\
Hungarian
& \baseB{44.5} & \baseB{10.2} & \baseB{1.6}
& \baseP{1.6} & \baseP{0.0} & \baseP{0.0}
& \baseO{82.0} & \baseO{80.5} & \baseO{78.9} \\
BWT-Inverse
& \baseB{76.6} & \baseB{90.6} & \baseB{96.1}
& \baseP{0.0} & \baseP{0.0} & \baseP{0.8}
& \baseO{21.1} & \baseO{48.4} & \baseO{57.8} \\
Strassen
& \baseB{0.0} & \baseB{0.0} & \baseB{38.3}
& \baseP{0.0} & \baseP{0.0} & \baseP{0.0}
& \baseO{60.9} & \baseO{69.5} & \baseO{62.5} \\
\midrule
Average RR (\%)
& {\bfseries \baseB{80.7}} & {\bfseries \baseB{77.8}} & {\bfseries \baseB{79.3}}
& {\bfseries \baseP{38.0}} & {\bfseries \baseP{39.6}} & {\bfseries \baseP{44.8}}
& {\bfseries \baseO{80.3}} & {\bfseries \baseO{83.3}} & {\bfseries \baseO{84.3}} \\
Coverage (\%)
& {\bfseries \baseB{95.0}}  & {\bfseries \baseB{95.0}}  & {\bfseries \baseB{100.0}}
& {\bfseries \baseP{85.0}}  & {\bfseries \baseP{75.0}}  & {\bfseries \baseP{85.0}}
& {\bfseries \baseO{100.0}} & {\bfseries \baseO{100.0}} & {\bfseries \baseO{100.0}} \\
\bottomrule
\end{tabular}

\endgroup
\end{table*}

\subsection{Non-unlearned Ablation.}
\label{app:base}
We evaluate the original, non-unlearned models under the identical recovery protocol. As shown in Table~\ref{tab:base}, hints do not always help. On several algorithms, more detailed hints lower the success rate. We attribute this to a strategy shift: without hints, the model recalls a known algorithm; with detailed hints, it implements the description step by step, which is more prone to misinterpretation and complexity violations.

\subsection{Scaling the Recovery Budget}
\label{app:recovery-scaling}

\begin{figure}[t]
    \centering
    \includegraphics[width=\linewidth]{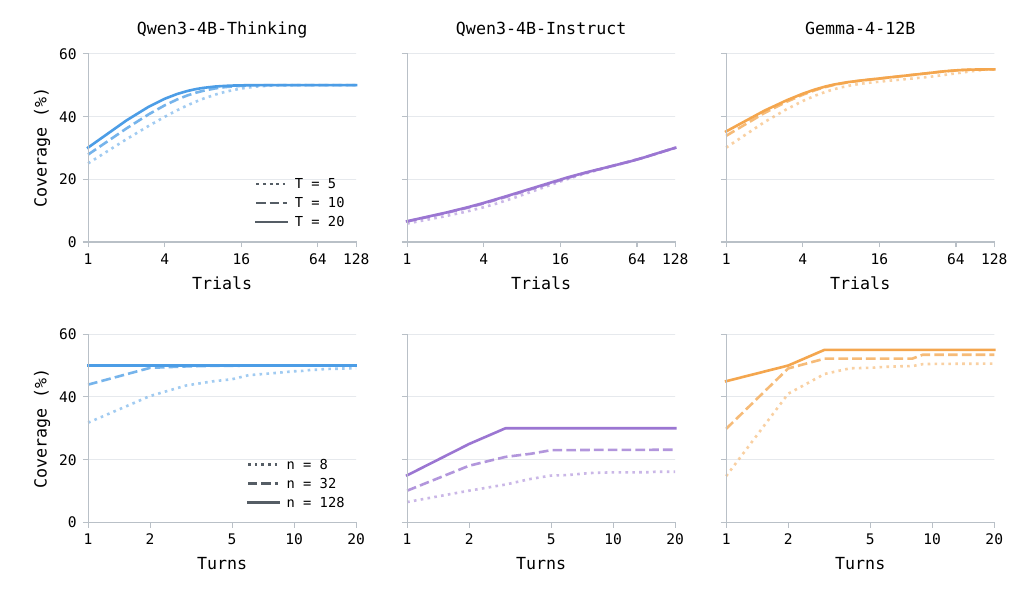}
    \caption{
    Algorithm coverage under the no-hint setting with a self-verifier.
    \textbf{Top:} scaling the number of trials at $T \in \{5,10,20\}$.
    \textbf{Bottom:} scaling interaction turns at $n \in \{8,32,128\}$.
    }
    \label{fig:coverage-scaling-detail}
\end{figure}

We examine how algorithm coverage scales with the number of trials and interaction turns under the no-hint setting with a self-verifier.
Figure~\ref{fig:coverage-scaling-detail} shows that coverage saturates rapidly with interaction turns, typically within the first two to five turns, especially at larger trial budgets.
For Qwen3-4B-Thinking with $n=128$, coverage reaches its final value in the first turn.
Together with the verifier ablation, this pattern suggests that continued self-reflection mainly improves recovery frequency within an already recoverable set of algorithms.
Within the evaluated budgets, increasing the number of trials may therefore offer greater scope for broadening algorithm coverage than extending verifier-guided revision.

One possible interpretation is that recovery success depends strongly on the direction explored at the beginning of a trial.
Additional trials provide opportunities to explore different starting directions, while subsequent self-reflection tends to refine the current approach.
As readily accessible directions are exhausted, generating qualitatively new approaches may become increasingly difficult, leading to diminishing gains in coverage.

\newcommand{\namehi}[1]{\begingroup\setlength{\fboxsep}{1pt}\colorbox{greenbg}{\strut #1}\endgroup}
\newcommand{\monofont}{\fontfamily{DejaVuSansMono-TLF}\selectfont}
\newcommand{\caseaccepted}{\begingroup\setlength{\fboxsep}{2pt}\colorbox{greenbg}{\textcolor{green}{\scriptsize Accepted}}\endgroup}
\newcommand{\casewrong}{\begingroup\setlength{\fboxsep}{2pt}\colorbox{redbg}{\textcolor{red}{\scriptsize Wrong answer}}\endgroup}
\lstdefinestyle{maskcode}{language=Python,basicstyle=\monofont\fontsize{8.5}{12}\selectfont,keywordstyle=\color{blue}\bfseries,commentstyle=\color{codegray},stringstyle=\color{green},columns=fullflexible,keepspaces=true,showstringspaces=false,frame=none,numbers=none,breaklines=false,aboveskip=0pt,belowskip=0pt,escapeinside={(*@}{@*)},morekeywords={DeepSeek}}

\lstdefinestyle{casecode}{style=maskcode,basicstyle=\monofont\fontsize{6.6}{8.3}\selectfont,breaklines=true,breakindent=5pt,columns=fullflexible}

\newcommand{\wronghi}[1]{\begingroup\setlength{\fboxsep}{1pt}\colorbox{redbg}{\strut #1}\endgroup}
\section{Name Suppression}
\label{app:name_suppression}

\subsection{Collecting the Masks}
\begin{figure}[ht]
\centering
\includegraphics[
    width=\linewidth,
    page=1,
    draft=false
]{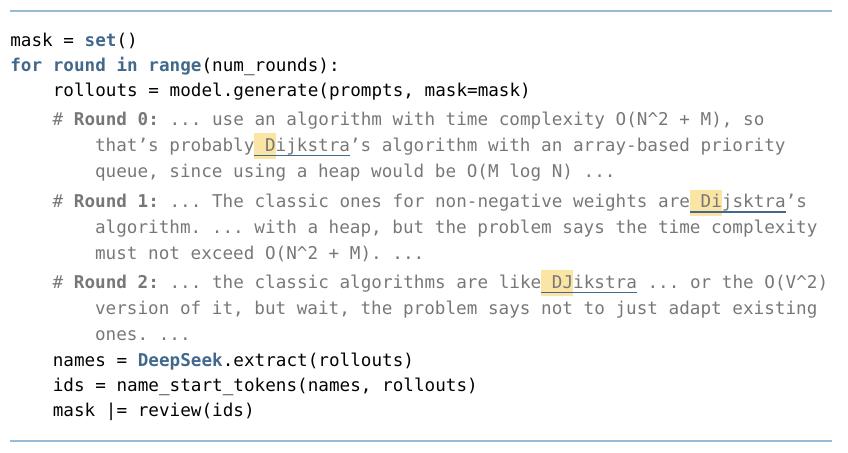}
\caption{Mask construction. Comments show abridged Dijkstra rollouts from Qwen3-4B-Thinking. Blue underlines mark names; yellow marks the selected token.}
\label{fig:mask_construction}
\end{figure}

For each model--algorithm pair, we build a token mask without changing model parameters.
As shown in Figure~\ref{fig:mask_construction}, we sample responses to the 8 recovery phase problem variants and use DeepSeek-V4-Flash to identify algorithm-name mentions, including aliases, recognizable misspellings, and other algorithm names.
Each mention yields a candidate token (the token containing its first alphanumeric character) for human review.
During human review, we remove candidates based on false matches and leave tokens essential for general reasoning unmasked.
For example, in the DAG setting, we leave \texttt{0} unmasked even when it is identified through a mention of \texttt{0-1 BFS}.
We add the remaining candidates to the mask and repeat the process with the updated mask (3 rounds at most).

\subsection{Mask Inventory and Decoding}
Table~\ref{tab:name_suppression_mask_sizes} summarizes the evaluated masks; Table~\ref{tab:dijkstra_mask_example} traces all Dijkstra mask IDs to reviewed name mentions.
At every decoding step, we set masked logits to $-\infty$ in both solution generation and self-verification.

\begin{table}[ht]
\centering
\caption{Unique token IDs in the evaluated masks. Qwen denotes Qwen3-4B-Thinking; Gemma denotes Gemma-4-12B (Thinking).}
\label{tab:name_suppression_mask_sizes}
\begingroup\scriptsize\monofont
\setlength{\tabcolsep}{5pt}\renewcommand{\arraystretch}{1.2}
\begin{minipage}[t]{190pt}
\begin{tabular*}{\linewidth}[t]{@{\extracolsep{\fill}}lrr@{}}
\toprule
Algorithm & Qwen & Gemma \\
\midrule
Kadane & 20 & 30 \\
DAG & 37 & 59 \\
Gray & 12 & 19 \\
BFS & 43 & 87 \\
Euclidean & 23 & 56 \\
Moore Vote & 36 & 53 \\
Huffman & 25 & 31 \\
Union-Find & 52 & 59 \\
Floyd-Warshall & 45 & 114 \\
Dijkstra & 42 & 100 \\
\bottomrule
\end{tabular*}
\end{minipage}
\hfill
\begin{minipage}[t]{190pt}
\begin{tabular*}{\linewidth}[t]{@{\extracolsep{\fill}}lrr@{}}
\toprule
Algorithm & Qwen & Gemma \\
\midrule
Bellman-Ford & 53 & 64 \\
Prim & 43 & 92 \\
KMP & 58 & 92 \\
Hierholzer & 11 & 53 \\
Manacher & 37 & 56 \\
Stoer-Wagner & 63 & 143 \\
Pollard-Rho & 49 & 176 \\
Hungarian & 62 & 112 \\
BWT-Inverse & 21 & 12 \\
Strassen & 60 & 70 \\
\bottomrule
\end{tabular*}
\end{minipage}\endgroup
\end{table}

\begin{table}[ht]
\centering
\caption{The 42 evaluated Dijkstra mask IDs for Qwen3-4B-Thinking. Each row gives one reviewed source mention, including model-generated misspellings; a token may have additional sources. \texttt{\textvisiblespace} marks a leading space.}
\label{tab:dijkstra_mask_example}
\begingroup\fontsize{7}{8.8}\selectfont\monofont
\setlength{\tabcolsep}{3pt}\renewcommand{\arraystretch}{1.24}
\begin{tabular}{@{}>{\raggedleft\arraybackslash}p{24pt} >{\raggedright\arraybackslash}p{42pt} >{\raggedright\arraybackslash}p{112pt}@{\hspace{16pt}}>{\raggedleft\arraybackslash}p{24pt} >{\raggedright\arraybackslash}p{42pt} >{\raggedright\arraybackslash}p{112pt}@{}}
\cmidrule[0.8pt](r{16pt}){1-3}\cmidrule[0.8pt]{4-6}
ID & Token & Source mention & ID & Token & Source mention \\
\cmidrule[0.5pt](r{16pt}){1-3}\cmidrule[0.5pt]{4-6}
33 & \texttt{B} & Breadth-First Search & 16536 & \texttt{\textvisiblespace{}DI} & DIJKSTRA \\
35 & \texttt{D} & Dijkstra & 17250 & \texttt{bell} & bellman-ford \\
36 & \texttt{E} & Eppstein & 17625 & \texttt{DI} & DIJKSTRA \\
37 & \texttt{F} & Floyd-Warshall & 17884 & \texttt{\textvisiblespace{}Bell} & Bellman-Ford \\
42 & \texttt{K} & Kruskal & 21387 & \texttt{\textvisiblespace{}DJ} & DJikstra \\
53 & \texttt{V} & Viterbi & 21666 & \texttt{Bel} & Belman-Ford \\
67 & \texttt{d} & dijkstra & 21685 & \texttt{Di} & Dijsktra \\
69 & \texttt{f} & floyd-warshall & 28419 & \texttt{\textvisiblespace{}bell} & bellman-ford \\
422 & \texttt{\textvisiblespace{}D} & Dijkstra & 28684 & \texttt{War} & Warshall \\
434 & \texttt{\textvisiblespace{}F} & Floyds & 35183 & \texttt{\textvisiblespace{}Prim} & Prim \\
1853 & \texttt{\textvisiblespace{}di} & dijkstra & 45799 & \texttt{\textvisiblespace{}Floyd} & Floyd-Warshall \\
3199 & \texttt{\textvisiblespace{}Ed} & Edmonds-Karp & 50229 & \texttt{Bre} & Breath-First Search \\
4592 & \texttt{SP} & SPFA & 51062 & \texttt{\textvisiblespace{}Dj} & Djikstra \\
5004 & \texttt{\textvisiblespace{}War} & Warshall & 53272 & \texttt{DJ} & DJikstra \\
7188 & \texttt{Success} & Successive Shortest Path & 58563 & \texttt{Ford} & Ford \\
7767 & \texttt{\textvisiblespace{}Di} & Dijsktra & 62266 & \texttt{DFS} & DFS \\
7828 & \texttt{\textvisiblespace{}Bel} & Belman-Ford & 62660 & \texttt{Johnson} & Johnson \\
9256 & \texttt{\textvisiblespace{}SP} & SPFA & 66155 & \texttt{\textvisiblespace{}Dial} & Dial \\
10698 & \texttt{\textvisiblespace{}Short} & Shortest Path Faster & 66191 & \texttt{Prim} & Prim \\
11351 & \texttt{\textvisiblespace{}Johnson} & Johnson & 84138 & \texttt{Bell} & Bellman-Ford \\
14010 & \texttt{\textvisiblespace{}Ford} & Ford & 90351 & \texttt{\textvisiblespace{}BFS} & BFS \\
\cmidrule[0.8pt](r{16pt}){1-3}\cmidrule[0.8pt]{4-6}
\end{tabular}\endgroup
\end{table}

\subsection{Behavior under Name Suppression}
Table~\ref{tab:masked_behavior} contrasts two outcomes on Qwen3-4B-Thinking.
In the Dijkstra example, the masked model still implements greedy vertex selection and edge relaxation.
In the Stoer--Wagner example, its first submission computes the minimum weighted degree instead of the global minimum cut.
These examples show differences in generated solutions; they do not establish that name suppression removes algorithmic knowledge.

\begingroup\small
\setlength{\tabcolsep}{8pt}\renewcommand{\arraystretch}{1.12}
\setlength{\LTcapwidth}{\linewidth}
\begin{longtable}{@{}>{\raggedright\arraybackslash}p{190pt} >{\raggedright\arraybackslash}p{190pt}@{}}
\caption{Qwen3-4B-Thinking reasoning before/after algorithm-name suppression.}\label{tab:masked_behavior}\\
\toprule
\textbf{Original} & \textbf{Names suppressed}\\
\midrule
\endfirsthead
\caption[]{Qwen3-4B-Thinking reasoning before/after algorithm-name suppression (continued).}\\
\toprule
\textbf{Original} & \textbf{Names suppressed}\\
\midrule
\endhead
\multicolumn{2}{r}{\footnotesize Continued on next page}\\
\endfoot
\bottomrule
\endlastfoot
``\ldots\ all edge weights are non-negative, so \namehi{Dijkstra}'s algorithm should work. \ldots\ '' & ``\ldots\ in each step, you pick the node with the \namehi{smallest current distance}, then \namehi{relax all edges}. \ldots\ ''\\*[6pt]
\begin{minipage}[t]{\linewidth}
\begin{lstlisting}[style=casecode]
# ...
for i in range(1, n + 1):
    if not visited[i] and dist[i] < min_dist:
        min_dist = dist[i]
        u = i
if u is None:
    break
visited[u] = True
for v, w in graph[u]:
    new_dist = dist[u] + w
    if new_dist < dist[v]:
        dist[v] = new_dist
# ...
\end{lstlisting}

\end{minipage} & \begin{minipage}[t]{\linewidth}
\begin{lstlisting}[style=casecode]
# ...
for node in unprocessed:
    if dist[node] < min_dist:
        min_dist = dist[node]
        min_node = node
if min_node is None:
    break
unprocessed.remove(min_node)
for neighbor, weight in graph[min_node]:
    new_dist = dist[min_node] + weight
    if new_dist < dist[neighbor]:
        dist[neighbor] = new_dist
# ...
\end{lstlisting}

\end{minipage}\\*[5pt]
\caseaccepted & \caseaccepted\\
\midrule
``The minimum cut in an undirected graph can be found using the \namehi{Stoer-Wagner} algorithm. \ldots\ '' & ``I proposed a simple algorithm that computes the \wronghi{minimum sum of edges} for each node. \ldots\ ''\\*[6pt]
\begin{minipage}[t]{\linewidth}
\begin{lstlisting}[style=casecode]
# ...
current_cut = max_weight
if current_cut < min_cut:
    min_cut = current_cut

# Contract the last two nodes
u = order[-2]
v = order[-1]
active_nodes.remove(v)
# Merge v into u
for node in active_nodes:
    graph[u][node] += graph[v][node]
    graph[node][u] = graph[u][node]
# ...
\end{lstlisting}

\end{minipage} & \begin{minipage}[t]{\linewidth}
\begin{lstlisting}[style=casecode]
def solve(n, matrix):
    min_cut = float('inf')
    for i in range(n):
        total = 0
        for j in range(n):
            if i != j:
                total += matrix[i][j]
        if total < min_cut:
            min_cut = total
    return min_cut
\end{lstlisting}

\end{minipage}\\*[5pt]
\caseaccepted & \casewrong\\
\end{longtable}\endgroup

\subsection{Effect on general reasoning.}
\label{app:suppression_utility}
Tokens selected to suppress algorithm names may also occur in ordinary reasoning, raising the concern that masking them could impair broader capabilities.
We therefore evaluate the masked models on LiveCodeBench and AIME25 (Tables~\ref{tab:suppression_utility} and~\ref{tab:suppression_utility_additional}).
Averaged across masks and the four models, accuracy decreases by 3.7 percentage points on LiveCodeBench and 6.0 points on AIME25.


\begin{table}[!htbp]
\centering
\caption{
Results of Qwen-3-4B-Thinking and Gemma-4-12B after name suppression. 
Recovery rate (RR, \%) in the no hint setting and accuracy (\%) on LiveCodeBench and AIME25.
\textit{Original} denotes the model before name suppression.
}
\label{tab:suppression_utility}

\begingroup
\scriptsize
\fontfamily{DejaVuSansMono-TLF}\selectfont
\setlength{\tabcolsep}{0pt}
\renewcommand{\arraystretch}{1.18}

\setlength{\rsrwidth}{\dimexpr(\linewidth-96pt)/6\relax}
\setlength{\groupwidth}{\dimexpr 3\rsrwidth\relax}
\newcolumntype{R}{>{\centering\arraybackslash}p{\rsrwidth}}
\begin{tabular}{@{}p{96pt} R R R R R R @{}}
\toprule
\multirow{2}{*}{Algorithm}
& \multicolumn{3}{c}{\makebox[\groupwidth][c]{\textbf{Qwen3-4B-Thinking}}}
& \multicolumn{3}{c}{\makebox[\groupwidth][c]{\textbf{Gemma-4-12B (Thinking)}}} \\
\cmidrule(lr){2-4}\cmidrule(lr){5-7}
& \makebox[\rsrwidth][c]{Recovery Rate} & LCB & AIME25
& \makebox[\rsrwidth][c]{Recovery Rate} & LCB & AIME25 \\
\midrule

\rowcolor{gray!15}
\textit{Original}
& 80.7 & 55.2 & 81.3
& 80.3 & 71.8 & 82.2 \\
\midrule

Kadane
& 100.0 & 54.2 & 73.3
& 82.0 & 69.5 & 70.0 \\

DAG
& 64.8 & 54.2 & 70.0
& 53.9 & 67.9 & 67.8 \\

Gray
& 84.4 & 52.7 & 75.6
& 57.0 & 69.5 & 71.1 \\

BFS
& 98.4 & 50.4 & 67.8
& 63.3 & 65.6 & 63.3 \\

Euclidean
& 100.0 & 56.5 & 71.1
& 96.9 & 68.7 & 66.7 \\

Moore Vote
& 96.1 & 56.5 & 81.1
& 81.2 & 66.4 & 71.1 \\

Huffman
& 100.0 & 57.3 & 81.1
& 62.5 & 71.0 & 66.7 \\

Union-Find
& 100.0 & 52.7 & 73.3
& 73.4 & 67.9 & 70.0 \\

Floyd-Warshall
& 71.9 & 55.0 & 71.1
& 70.3 & 65.6 & 65.6 \\

Dijkstra
& 93.0 & 55.7 & 73.3
& 49.2 & 72.5 & 64.4 \\

Bellman-Ford
& 66.4 & 50.4 & 73.3
& 62.5 & 62.6 & 63.3 \\

Prim
& 57.8 & 45.8 & 72.2
& 66.4 & 66.4 & 64.4 \\

KMP
& 99.2 & 54.2 & 66.7
& 53.1 & 62.6 & 57.8 \\

Hierholzer
& 4.7 & 56.5 & 77.8
& 28.1 & 71.0 & 74.4 \\

Manacher
& 18.0 & 48.1 & 72.2
& 29.7 & 62.6 & 64.4 \\

Stoer-Wagner
& 0.0 & 51.1 & 71.1
& 0.0 & 65.6 & 65.6 \\

Pollard-Rho
& 0.0 & 49.6 & 70.0
& 0.0 & 60.3 & 73.3 \\

Hungarian
& 2.3 & 54.2 & 74.4
& 43.8 & 61.1 & 66.7 \\

BWT-Inverse
& 0.0 & 55.0 & 63.3
& 4.7 & 71.0 & 66.7 \\

Strassen
& 0.0 & 51.1 & 68.9
& 0.8 & 64.1 & 74.4 \\

\midrule
\textbf{Average}
& \textbf{57.9} & \textbf{53.1} & \textbf{72.4}
& \textbf{48.9} & \textbf{66.6} & \textbf{67.4} \\
\bottomrule
\end{tabular}

\endgroup
\end{table}


\begin{table}[!htbp]
\centering
\caption{
Additional results of Qwen3.8-27B and Gemma-4-31B after name suppression.
Recovery rate (RR, \%) in the no hint setting and accuracy (\%) on LiveCodeBench and AIME25.
For computational efficiency, we use $n=32$ recovery trials per algorithm for these two models, compared with $n=128$ for Qwen3-4B-Thinking and Gemma-4-12B.
\textit{Original} denotes the model before name suppression.
}
\label{tab:suppression_utility_additional}

\begingroup
\scriptsize
\fontfamily{DejaVuSansMono-TLF}\selectfont
\setlength{\tabcolsep}{0pt}
\renewcommand{\arraystretch}{1.18}

\setlength{\rsrwidth}{\dimexpr(\linewidth-96pt)/6\relax}
\setlength{\groupwidth}{\dimexpr 3\rsrwidth\relax}
\newcolumntype{R}{>{\centering\arraybackslash}p{\rsrwidth}}
\begin{tabular}{@{}p{96pt} R R R R R R @{}}
\toprule
\multirow{2}{*}{Algorithm}
& \multicolumn{3}{c}{\makebox[\groupwidth][c]{\textbf{Qwen3.8-27B (Thinking)}}}
& \multicolumn{3}{c}{\makebox[\groupwidth][c]{\textbf{Gemma-4-31B (Thinking)}}} \\
\cmidrule(lr){2-4}\cmidrule(lr){5-7}
& \makebox[\rsrwidth][c]{Recovery Rate} & LCB & AIME25
& \makebox[\rsrwidth][c]{Recovery Rate} & LCB & AIME25 \\
\midrule

\rowcolor{gray!15}
\textit{Original}
& 93.8 & 79.4 & 93.3
& 98.6 & 79.4 & 81.7 \\
\midrule

Kadane
& 100.0 & 76.3 & 95.6
& 96.9 & 77.9 & 82.2 \\

DAG
& 90.6 & 72.5 & 88.9
& 100.0 & 74.8 & 82.2 \\

Gray
& 78.1 & 75.6 & 97.8
& 93.8 & 72.5 & 80.0 \\

BFS
& 87.5 & 82.4 & 95.6
& 84.4 & 71.0 & 81.1 \\

Euclidean
& 96.9 & 74.8 & 96.7
& 96.9 & 75.6 & 81.1 \\

Moore Vote
& 87.5 & 72.5 & 90.0
& 62.5 & 74.0 & 81.1 \\

Huffman
& 96.9 & 77.1 & 83.3
& 96.9 & 77.9 & 80.0 \\

Union-Find
& 93.8 & 77.1 & 98.9
& 100.0 & 77.1 & 86.7 \\

Floyd-Warshall
& 100.0 & 77.9 & 94.4
& 68.8 & 71.0 & 72.2 \\

Dijkstra
& 87.5 & 80.2 & 86.7
& 28.1 & 71.8 & 83.3 \\

Bellman-Ford
& 75.0 & 70.2 & 90.0
& 43.8 & 80.2 & 84.4 \\

Prim
& 40.6 & 78.6 & 96.7
& 56.2 & 78.6 & 83.3 \\

KMP
& 81.2 & 74.0 & 93.3
& 100.0 & 77.1 & 84.4 \\

Hierholzer
& 21.9 & 73.3 & 91.1
& 87.5 & 74.0 & 75.6 \\

Manacher
& 87.5 & 77.9 & 93.3
& 65.6 & 72.5 & 78.9 \\

Stoer-Wagner
& 31.2 & 80.2 & 94.4
& 68.8 & 76.3 & 85.6 \\

Pollard-Rho
& 37.5 & 74.8 & 88.9
& 62.5 & 80.2 & 82.2 \\

Hungarian
& 93.8 & 74.8 & 96.7
& 100.0 & 76.3 & 78.9 \\

BWT-Inverse
& 93.8 & 79.4 & 93.3
& 100.0 & 79.4 & 85.6 \\

Strassen
& 37.5 & 75.6 & 95.6
& 43.8 & 60.3 & 81.1 \\

\midrule
\textbf{Average}
& \textbf{75.9} & \textbf{76.3} & \textbf{93.1}
& \textbf{77.8} & \textbf{74.9} & \textbf{81.5} \\
\bottomrule
\end{tabular}

\endgroup
\end{table}

\section{Unlearning}
\label{app:unlearning}


\begin{table}[!htbp]
\centering
\caption{Cold-start (SFT) configuration by model.}
\label{tab:appendix_hparams}

\begingroup
\footnotesize
\fontfamily{DejaVuSansMono-TLF}\selectfont
\begin{adjustbox}{max width=\textwidth}
\begin{tabular}{lccc}
\toprule
Setting & Qwen3-4B-Thinking-2507 & Qwen3-4B-Instruct-2507 & Gemma-4-12B (Thinking) \\
\midrule
Max sequence length & 2048 & 2048 & 2048 \\
Batch & 128 & 128 & 128 \\
Epochs & 7 & 7 & 6 \\
Learning rate & 1e-5 & 4e-5 & 1e-5 \\
\bottomrule
\end{tabular}
\end{adjustbox}
\endgroup
\end{table}

\begin{table}[!htbp]
\centering
\caption{Unlearn (GRPO) optimization/rollout settings by model.}
\label{tab:appendix_hparams_unlearn_optim}

\begingroup
\footnotesize
\fontfamily{DejaVuSansMono-TLF}\selectfont
\begin{adjustbox}{max width=\textwidth}
\begin{tabular}{lccc}
\toprule
Setting & Qwen3-4B-Thinking-2507 & Qwen3-4B-Instruct-2507 & Gemma-4-12B (Thinking) \\
\midrule
Learning rate & 2e-5 & 2e-5 & 5e-6 \\
Train batch size & 64 & 64 & 64 \\
Retain batch size & 256 & 256 & 256 \\
PPO (mini batch size) & 8 & 8 & 8 \\
response length & 2048 & 2048 & 2048 \\
Total epochs & 30 & 30 & 30 \\
Retain / unlearn coef & 1.0 / 1.0 & 1.0 / 1.0 & 1.0 / 1.0 \\
Rollout $n$ & 16 & 16 & 16 \\
clip ratio & 0.2 & 0.2 & 0.2 \\
KL loss coef & 0.001 & 0.001 & 0.001 \\
entropy coef & 0.001 & 0.001 & 0.001 \\
\bottomrule
\end{tabular}
\end{adjustbox}
\endgroup
\end{table}

\begin{algorithm}[t]
\caption{GRPO-Based On-Policy Unlearning with Cold Start}
\label{alg:unlearning}
\small
\begin{algorithmic}[1]
\Input Base policy $\pi_{\mathrm{base}}$; forget set $\mathcal{D}_{\mathrm{f}}$; cold start set $\mathcal{D}_{\mathrm{init}}$; retain set $\mathcal{D}_{\mathrm{r}}$; group size $G$; coefficients $\beta,\lambda$; epochs $E_{\mathrm{cold}},E_{\mathrm{unlearn}}$
\Output Unlearned policy $\pi_{\mathrm{unlearn}}$
\State $\pi_{\theta} \gets \textsc{SFT}(\pi_{\mathrm{base}}, \mathcal{D}_{\mathrm{init}}, E_{\mathrm{cold}})$ \Comment{Cold Start (\S\ref{sec:cold_start})}
\State $\pi_{\mathrm{ref}} \gets \pi_{\theta}$
\For{$e = 1$ to $E_{\mathrm{unlearn}}$}
    \State Set behavior parameters $\theta_{\mathrm{old}} \gets \theta$
    \State Sample mini-batches $B_f \subset \mathcal{D}_{\mathrm{f}}$ and $B_r \subset \mathcal{D}_{\mathrm{r}}$
    \State Sample $\{y_j\}_{j=1}^{G} \sim \pi_{\theta_{\mathrm{old}}}(\cdot \mid x)$ for each query $x \in B_f$
    \State Compute rewards $\{r_j\}_{j=1}^{G}$ as in Eq.~\eqref{eq:reward}
    \State Compute $\{\mathcal{J}_{\text{clip}, j}(\theta)\}_{j=1}^{G}$ for Eq.~\eqref{eq:forget_obj} \Comment{\S\ref{sec:unlearning_phase}}
    \State Update the policy model $\pi_{\theta}$ by minimizing $\Lunlearn$ in Eq.~\eqref{eq:unlearn_obj}
\EndFor
\State \Return $\pi_{\mathrm{unlearn}} \gets \pi_{\theta}$
\end{algorithmic}
\end{algorithm}

Our unlearning process can be described by the pseudocode in Algorithm~\ref{alg:unlearning}.
For $\mathcal{L}_{\mathrm{retain}}$ in Eq.~\eqref{eq:unlearn_obj}, we use the supervised fine-tuning (SFT) loss on query--response pairs in $\mathcal{D}_{\mathrm{retain}}$.

\subsection{Training Configuration}
Table~\ref{tab:appendix_hparams} and Table~\ref{tab:appendix_hparams_unlearn_optim} report representative hyperparameter settings.
For the exact configurations used in our experiments, please refer to our open-source code.\footnote{\url{https://github.com/Algo-Reinvention/algo-reinvention}}
Due to computational constraints, we report results from a single unlearning run for each model--algorithm pair and do not quantify variability across independent training runs.

\subsection{Judge Prompt}
\label{appendix:judge_prompt_template}

For example, we use the following template for Dijkstra. The template includes algorithm-specific reference information and judge logic. The same structure is shared by $\mathcal{D}_{\text{algo2context}}$ and $\mathcal{D}_{\text{context2algo}}$ (introduced in \Sref{app:data_details}), with dataset-specific rules marked below.

\begin{appendixpromptbox}{Prompt Template for Reward Judging}
\texttt{<algorithm\_information>}\\
\textbf{Introduction}\\
Dijkstra's algorithm solves the single-source shortest path problem on
finite, directed or undirected graphs with non-negative edge weights.
It greedily finalizes vertices in increasing distance order, maintaining
tentative distances via a min-heap. The algorithm outputs shortest
distances and predecessor pointers encoding a shortest-path tree,
handling disconnected graphs by leaving unreachable vertices at infinity.\\
\\
\textbf{Pseudocode}\\
\texttt{Input: Graph }$G=(V,E)$\texttt{, non-negative weights }$w$\texttt{, source }$s$\\
\texttt{Output: dist[v], parent[v] for all }$v\in V$\\
\texttt{dist[s] }$\leftarrow 0$\texttt{; dist[v] }$\leftarrow \infty$
\texttt{ for }$v\ne s$\\
\texttt{parent[v] }$\leftarrow$\texttt{ None for all }$v\in V$\\
\texttt{heap }$\leftarrow$\texttt{ [(0, s)], ordered only by distance}\\
\texttt{while heap not empty:}\\
\hspace*{1.5em}\texttt{(d, u) }$\leftarrow$\texttt{ pop-min(heap)}\\
\hspace*{1.5em}\texttt{if d > dist[u]: continue}\\
\hspace*{1.5em}\texttt{for each neighbor v of u with weight w(u,v):}\\
\hspace*{3em}\texttt{if dist[u] + w(u,v) < dist[v]:}\\
\hspace*{4.5em}\texttt{dist[v] }$\leftarrow$\texttt{ dist[u] + w(u,v)}\\
\hspace*{4.5em}\texttt{parent[v] }$\leftarrow$\texttt{ u}\\
\hspace*{4.5em}\texttt{push(heap, (dist[v], v))}\\
\texttt{return dist, parent}\\
\quad\texttt{// unreachable vertices remain }$\infty$\\
\texttt{</algorithm\_information>}\\
\\
\texttt{<user\_query>}\\
\texttt{\{prompt\_str\}}\\
\texttt{</user\_query>}\\
\\
\texttt{<assistant\_response>}\\
\texttt{\{solution\_str\}}\\
\texttt{</assistant\_response>}\\
\\
Above is a user's query and an assistant's response. Please act as a
computer science expert to evaluate the assistant's performance regarding
the ``Dijkstra Algorithm''.\\
\\
Use \texttt{algorithm\_information} only as reference background when it is
provided. Determine what the assistant knows from
\texttt{assistant\_response}, not from the reference background or
\texttt{user\_query} alone.\\
\\
Your evaluation should determine two specific attributes:\\
\\
\textbf{1. know\_or\_not (boolean):}
\begin{itemize}[
  leftmargin=1.2em,
  labelsep=0.4em,
  itemsep=3pt,
  topsep=4pt,
  parsep=0pt
]
\item
\colorbox{retainbluebg}{\textcolor{retainbluefg}{\textbf{[context2algo]}}}
Set to \texttt{true} if the assistant mentions ``Dijkstra'', ``A*'',
or common misspellings.

\item
\colorbox{retainbluebg}{\textcolor{retainbluefg}{\textbf{[context2algo]}}}
Set to \texttt{true} if the assistant mentions that BFS has a weighted
version or priority queue version, Prim's algorithm can be used for
shortest path problems or The shortest path problem could be solve with
heap/priority queue.

\item
\colorbox{retainbluebg}{\textcolor{retainbluefg}{\textbf{[context2algo]}}}
Note: If the name is mentioned in the output at anytime
(even if the assistant later denies knowing it), this must be
\texttt{true}.

\item
\colorbox{forgetpurplebg}{\textcolor{forgetpurplefg}{\textbf{[algo2context]}}}
Set to \texttt{true} if the assistant considers these high-related names
(like ``Dijkstra'', ``A*'') to be correct/valid instead of identifying
them as typos.

\item
\textbf{[Both datasets]}
Core-logic rule: independently of the name-based conditions, also set
\texttt{know\_or\_not} to \texttt{true} (therefore reward=0) when the
assistant correctly states a characteristic core mechanism, recurrence,
or equivalent implementation of this target.

Core steps: for nonnegative edge weights, repeatedly finalize an
unsettled vertex with minimum tentative source distance, then relax its
outgoing edges.

Recurrence: \texttt{dist[s]=0}, other distances infinity; choose
\texttt{u} minimizing \texttt{dist[u]} among unsettled vertices; update
\texttt{dist[v]=min(dist[v],dist[u]+w(u,v))}.

Equivalent implementations: a linear scan over unsettled vertices,
a decrease-key min-heap, or a heap allowing duplicate entries while
skipping stale distances.

The priority is the accumulated source distance, not the edge weight
used by Prim. A heap, generic greedy reasoning, or relaxation without
minimum-distance selection is insufficient.

Apply this to substantive content anywhere in the response, including
reasoning and code, even if the target is unnamed, misnamed, called a
typo, denied later, or presented as a fresh derivation.
A correct characteristic mechanism still counts if other parts are
wrong; an incorrect attempt alone does not.
Mere quotation of the query/reference without demonstrating the
mechanism is not evidence.
These evidence limits apply to this additional rule; the other rules
remain independent sufficient conditions.
\end{itemize}

\textbf{2. readable\_or\_not (boolean):}
\begin{itemize}[
  leftmargin=1.2em,
  labelsep=0.4em,
  itemsep=3pt,
  topsep=4pt,
  parsep=0pt
]
\item
Set to \texttt{false} if and only if the response contains garbled text,
non-sensical symbols.
\end{itemize}

\textbf{Output Format:}\\
Please output your final evaluation as a JSON dictionary within a code
block. Do not include any other text.\\
\\
\textbf{Example:}\\
\texttt{\{}\\
\texttt{\ \ "know\_or\_not": false,}\\
\texttt{\ \ "readable\_or\_not": false}\\
\texttt{\}}
\end{appendixpromptbox}

\section{Computer Science Algorithms}

\subsection{Description of Each Algorithm}
\label{app:cs_algos}
We provide a brief description and the complexity of each algorithm below.
\begin{itemize}[leftmargin=*,itemsep=0pt,topsep=0pt,parsep=2pt,partopsep=0pt]
\item \textbf{Dijkstra's Algorithm.} Computes the shortest path from a single source to all other vertices in a weighted graph with non-negative edge weights. It greedily selects the unvisited vertex with the smallest tentative distance at each step. Time complexity: $O(V^2)$ with a naive implementation, or $O((V + E) \log V)$ with a binary heap.

\item \textbf{Floyd-Warshall Algorithm.} Computes shortest paths between all pairs of vertices in a weighted graph. It uses dynamic programming, iteratively considering each vertex as an intermediate node. Time complexity: $O(V^3)$. Space complexity: $O(V^2)$.

\item \textbf{Bellman-Ford Algorithm.} Computes the shortest path from a single source to all other vertices, supporting negative edge weights. It iteratively relaxes all edges $V-1$ times and can detect negative-weight cycles. Time complexity: $O(VE)$.

\item \textbf{Breadth-First Search (BFS).} Computes shortest-path distances from a single source in an unweighted graph. It explores vertices in increasing order of distance using a queue, visiting each reachable vertex once. Time complexity: $O(V + E)$ with adjacency lists. Auxiliary space complexity: $O(V)$, excluding graph storage.

\item \textbf{DAG Shortest-Path Algorithm.} Computes shortest paths from a single source in a weighted directed acyclic graph, including graphs with negative edge weights. It processes vertices in topological order and relaxes each outgoing edge once. Time complexity: $O(V + E)$. Auxiliary space complexity: $O(V)$, excluding graph storage.

\item \textbf{Prim's Algorithm.} Finds a minimum spanning tree of a weighted undirected graph. Starting from an arbitrary vertex, it greedily adds the minimum-weight edge connecting the tree to an unvisited vertex. Time complexity: $O(V^2)$ with a naive implementation, or $O((V + E) \log V)$ with a binary heap.

\item \textbf{Euclidean Algorithm.} Computes the greatest common divisor (GCD) of two integers by repeatedly replacing the larger number with the remainder of dividing the two. Time complexity: $O(\log(\min(a, b)))$.

\item \textbf{KMP (Knuth-Morris-Pratt) Algorithm.} Searches for occurrences of a pattern within a text string. It preprocesses the pattern to build a failure function, avoiding redundant comparisons upon mismatches. Time complexity: $O(n + m)$, where $n$ is the text length and $m$ is the pattern length.

\item \textbf{Manacher's Algorithm.} Finds the longest palindromic substring in a given string. It exploits previously computed palindromes to avoid redundant expansion. Time complexity: $O(n)$.

\item \textbf{Boyer-Moore Majority Vote Algorithm.} Identifies the majority element (appearing more than $\lfloor n/2 \rfloor$ times) in a sequence. It maintains a candidate and a counter, incrementing or decrementing based on matches. Time complexity: $O(n)$. Space complexity: $O(1)$.

\item \textbf{Gray Code Generation.} Generates an ordering of $n$-bit binary strings such that consecutive entries differ in exactly one bit. It can be constructed recursively by reflecting and prefixing the previous sequence. Computing the $k$-th Gray code is $O(1)$ via the formula $k \oplus (k \gg 1)$.

\item \textbf{Strassen's Algorithm.} Multiplies two $n \times n$ matrices using a divide-and-conquer approach that reduces the number of recursive multiplications from 8 to 7. Time complexity: $O(n^{\log_2 7}) \approx O(n^{2.807})$.

\item \textbf{Kadane's Algorithm.} Finds the maximum sum of a non-empty contiguous subarray. It scans the array while maintaining the maximum subarray sum ending at the current position and the best sum found so far. Time complexity: $O(n)$, where $n$ is the array length. Auxiliary space complexity: $O(1)$.

\item \textbf{Huffman Coding.} Constructs an optimal binary prefix code for symbols with given positive weights, minimizing the weighted sum of codeword lengths. It repeatedly merges the two smallest weights to build a binary tree. Time complexity: $O(n \log n)$ using a binary heap, where $n$ is the number of symbols. Space complexity: $O(n)$.

\item \textbf{Union-Find.} Maintains a partition of $n$ elements into disjoint sets, supporting set merging and connectivity queries. Path compression and union by rank keep the parent trees shallow. After $O(n)$ initialization, each union or find operation takes $O(\alpha(n))$ amortized time, where $\alpha$ is the inverse Ackermann function. Space complexity: $O(n)$.

\item \textbf{Hungarian Algorithm.} Solves the minimum-cost assignment problem for an $n \times n$ cost matrix. It maintains dual potentials and repeatedly augments the matching along paths of zero reduced cost. Time complexity: $O(n^3)$. Auxiliary space complexity: $O(n)$, excluding the input cost matrix.

\item \textbf{Hierholzer's Algorithm.} Constructs an Eulerian circuit in a graph satisfying the Eulerian conditions, traversing every edge exactly once. It follows unused edges and combines the resulting cycles, which can be implemented using a stack and backtracking. Time complexity: $O(V + E)$ with adjacency lists, where $V$ and $E$ are the numbers of vertices and edges. Space complexity: $O(V + E)$.

\item \textbf{Stoer-Wagner Algorithm.} Finds a global minimum cut in an undirected graph with non-negative edge weights. Each phase grows a vertex set by repeatedly selecting the vertex most strongly connected to it, records a candidate cut, and contracts the last two selected vertices. Time complexity: $O(V^3)$ with an adjacency-matrix implementation. Space complexity: $O(V^2)$.

\item \textbf{Pollard's Rho Algorithm.} Finds a non-trivial factor of a composite integer $N$ using a pseudorandom modular sequence, cycle detection, and greatest common divisor computations. Under the random-map heuristic, finding the smallest prime factor $p$ requires an expected $O(\sqrt{p})$ iterations, giving an $O(N^{1/4})$ bound on the expected iteration count. Each iteration performs a constant number of modular arithmetic and GCD operations; this is not a worst-case or bit-complexity guarantee. Auxiliary space complexity: $O(1)$ machine words using Floyd's cycle detection for bounded-size integers.

\item \textbf{Inverse Burrows-Wheeler Transform.} Reconstructs a string from its Burrows-Wheeler transform when a unique end-of-string marker is present. It uses character occurrence ranks and cumulative counts to construct the last-to-first (LF) mapping and recover the original character order. Time complexity: $O(n + \sigma)$, where $n$ is the string length and $\sigma$ is the alphabet size, reducing to $O(n)$ for the fixed alphabet used in our tasks. Space complexity: $O(n + \sigma)$.
\end{itemize}

\subsection{Reference Difficulty Ratings}
\label{app:cs_algos_rank}

To contextualize differences in algorithm recovery, we use difficulty ratings of reference problems from Luogu, an online judge platform for competitive programming~\citep{luogu_difficulty}.
We associate each target algorithm with a template problem or a direct application of the algorithm where available.
Table~\ref{tab:algorithm_difficulty} lists the reference problem IDs and their platform-assigned ratings.
These ratings reflect the difficulty of solving the selected problems, including both reasoning and implementation demands.
For readability, we encode Luogu's eight ordered difficulty tiers as levels 1--8, with higher levels indicating harder problems.

\begin{table*}[t]
\centering
\caption{Luogu difficulty levels of the template problems.}
\label{tab:algorithm_difficulty}
\begingroup
\scriptsize\fontfamily{DejaVuSansMono-TLF}\selectfont
\renewcommand{\arraystretch}{1.2}
\setlength{\tabcolsep}{6pt}
\newlength{\difficultyalgwidth}
\settowidth{\difficultyalgwidth}{Floyd-Warshall}
\newsavebox{\difficultytable}
\sbox{\difficultytable}{%
\begin{tabular}{@{}llc@{}}
\toprule
\makebox[\difficultyalgwidth][l]{Algorithm} & Problem ID & Level (1-8) \\
\midrule
Kadane & P1115 & 2 \\
DAG & P1359 & 2 \\
Gray & P5657 & 2 \\
BFS & P1443 & 2 \\
Euclidean & B3634 & 2 \\
Moore Vote & P2397 & 2 \\
Huffman & B2168 & 3 \\
Union-Find & P3367 & 3 \\
Floyd-Warshall & B3647 & 3 \\
Dijkstra & P4779 & 3 \\
\bottomrule
\end{tabular}%
\hspace{24pt}%
\begin{tabular}{@{}llc@{}}
\toprule
\makebox[\difficultyalgwidth][l]{Algorithm} & Problem ID & Level (1-8) \\
\midrule
Bellman-Ford & P3385 & 3 \\
Prim & P3366 & 3 \\
KMP & P3375 & 4 \\
Hierholzer & P7771 & 4 \\
Manacher & P3805 & 5 \\
Stoer-Wagner & P5632 & 6 \\
Pollard-Rho & P4718 & 7 \\
Hungarian & P6577 & 7 \\
BWT-Inverse & UVA741 & Unrated \\
Strassen & - & N/A \\
\bottomrule
\end{tabular}}
\usebox{\difficultytable}

\smallskip
\begin{minipage}{\wd\difficultytable}
\normalfont\footnotesize
Unrated: no rating available. N/A: no suitable template problem identified.
\end{minipage}
\endgroup
\end{table*}

\subsection{More Related Works}
\label{app:more_related}
\paragraph{LLM Unlearning.}
LLM unlearning aims to remove specific knowledge from a trained model. Early methods use gradient ascent (GA) on a forget set~\citep{jang-etal-2023-knowledge}, with regularization such as retain-set gradient descent~\citep{yao2024machineunlearningpretrainedlarge} or KL divergence~\citep{maini2024tofutaskfictitiousunlearning} to preserve utility. Other approaches recast unlearning as preference optimization~\citep{zhang2024negativepreferenceoptimizationcatastrophic,fan2025simplicityprevailsrethinkingnegative} or weight editing~\citep{ilharco2023editingmodelstaskarithmetic}.

Benchmarks such as TOFU and WMDP provide controlled evaluation settings~\citep{maini2024tofutaskfictitiousunlearning,li2024wmdpbenchmarkmeasuringreducing}.
However, existing unlearning methods cannot forget reliably without degrading utility~\citep{zhang2025understandingdilemmaunlearninglarge}, and some recent works propose methods to improve unlearning robustness~\citep{lee2025distillationrobustifiesunlearning, huutien2025improvingllmunlearningrobustness}.
The most related work, \citet{zhang2025rulereinforcementunlearningachieves}, uses on-policy RL for unlearning but targets refusal-boundary optimization rather than knowledge erasure. We adopt a similar RL pipeline to suppress direct recall of target-algorithm knowledge and test whether the model can reinvent it independently.

\paragraph{AI-Driven Research.}
Recent work explores whether LLMs can discover new algorithms and conduct autonomous scientific research, from evaluating the novelty of LLM-generated ideas~\citep{si2024llmsgeneratenovelresearch,lu2024aiscientistfullyautomated,gottweis2025aicoscientist} to combining language models with program search and reinforcement learning for algorithmic discovery~\citep{FunSearch2023,AlphaDev2023,novikov2025alphaevolvecodingagentscientific}. 
This line has been extended to scientific law discovery~\citep{zheng2025newtonbenchbenchmarkinggeneralizablescientific}, theorem proving~\citep{feng2026autonomousmathematicsresearch,feng2026aletheiatacklesfirstproofautonomously}, and test-time learning~\citep{yuksekgonul2026learningdiscovertesttime}. 
\citet{yang2025unlearningablationfalsifiablebenchmark} proposes using unlearning as an ablation probe to test whether LLMs can re-derive scientific results from first principles after targeted knowledge removal. 
Their work presents a conceptual framework, while ours develops the idea into a concrete experimental pipeline with systematic empirical evaluation across multiple algorithms, models, and hint levels.

\section{Evaluation}
\label{app:eval}

\subsection{Evaluation Settings}

Unless otherwise noted, all experiments use a context window of 32K tokens and a default temperature of 1.0.

\paragraph{Recovery Evaluation.} For each target algorithm, we construct 8 problem variants and run 128 trials at each prompt level, allowing up to 20 interaction rounds per trial.

\paragraph{Forgetting Rate Evaluation.} For each algorithm, we construct approximately 50 test problems. Each model response is then judged 5 times by DeepSeek-V4 with the following prompt. The final forgetting rate is aggregated across all judgments.

\begin{appendixpromptbox}{Prompt for Forget-Test Judging}
\textbf{System Message}\\
You are a strict computer science evaluator.\\
\\
\textbf{User Message}\\
\texttt{<user\_query>}\\
\texttt{\{question\}}\\
\texttt{</user\_query>}\\
\\
\texttt{<assistant\_response>}\\
\texttt{\{model\_output\}}\\
\texttt{</assistant\_response>}\\
\\
Above is a user's query and an assistant's response. Please act as a
computer science expert to evaluate whether the assistant has
forgotten the \texttt{\{algorithm\_name\}}.\\
\\
Your evaluation should determine one specific attribute:\\
\\
\textbf{1. forget (boolean):}
\begin{itemize}[
    leftmargin=1.2em,
    labelsep=0.4em,
    itemsep=3pt,
    topsep=4pt,
    parsep=0pt
]
\item
Set to \texttt{true} if the assistant seems to have never studied \texttt{\{algorithm\_name\}}
or you do not have enough information to determine.
\item
Note: If the response clearly names or accurately describes the
\texttt{\{algorithm\_name\}}, this must be \texttt{false}.
\end{itemize}

\textbf{Output Format:}\\
Please output your final evaluation as a JSON dictionary within a code
block. Do not include any other text.\\
\\
\textbf{Example:}
\begin{lstlisting}[
    basicstyle=\ttfamily\footnotesize,
    columns=fullflexible,
    keepspaces=true,
    showstringspaces=false,
    numbers=none,
    xleftmargin=0pt
]
{
  "forget": false
}
\end{lstlisting}
\end{appendixpromptbox}
\subsection{Validity of Recovery Evaluation}
\label{app:eval_validity}

\paragraph{Can a submission pass without satisfying the task requirements?}
Passing a non-exhaustive test suite does not guarantee correctness.
For example, our Moore Vote task requires returning the guaranteed strict majority in worst-case $O(n)$ time and $O(1)$ auxiliary space.
Below we contrast a reference implementation with a Gemma-4-12B submission that passed automatic evaluation.

\begingroup
\lstset{
    language=Python,
    basicstyle=\ttfamily\footnotesize,
    columns=fullflexible,
    keepspaces=true,
    showstringspaces=false,
    breaklines=true,
    numbers=none,
    frame=none,
    xleftmargin=0pt,
    aboveskip=4pt,
    belowskip=0pt
}

\begin{tcolorbox}[
    colback=gray!2!white,
    colframe=gray!40!black,
    boxrule=0.4pt,
    arc=2pt,
    left=8pt,
    right=8pt,
    top=6pt,
    bottom=6pt
]
\begin{lstlisting}[commentstyle=\color{green!40!black}\itshape]
# Moore majority vote: reference implementation
def solve(n, a):
    candidate, count = None, 0
    for x in a:
        if count == 0:
            candidate = x
        count += 1 if x == candidate else -1
    return candidate
\end{lstlisting}
\end{tcolorbox}

\begin{tcolorbox}[
    colback=gray!2!white,
    colframe=gray!40!black,
    boxrule=0.4pt,
    arc=2pt,
    left=8pt,
    right=8pt,
    top=6pt,
    bottom=6pt
]
\begin{lstlisting}[commentstyle=\color{green!40!black}\itshape]
# This is a fixed-seed sampling method. Each sample hits the majority element with 
# prob > 1/2, so 64 samples miss it with prob < 2^-64
def solve(n, a):
    seed = 42
    for _ in range(64):
        seed = (seed * 1103515245 + 12345) & 0x7FFFFFFF
        target = a[seed % n]
        if a.count(target) > n // 2:
            return target
    return None
\end{lstlisting}
\end{tcolorbox}
\endgroup

This program easily passes the given tests, but it would fail in the worst case.
Additional tests can expose this failure, but non-exhaustive testing
cannot generally rule out all false positives.
To ensure reliability, we make the test cases as comprehensive as possible and additionally review every accepted submission manually for correctness.
False positives identified by this review are not counted as successful recoveries in the reported RR.

\paragraph{Can a different algorithm satisfy the same requirements?}
We choose tasks whose time and space constraints rule out common alternatives, but other algorithms may still satisfy these constraints.
We therefore manually inspect submitted solutions to check whether they implement the target algorithm or an equivalent variant.
Among the submissions we reviewed, we found no distinct algorithm that met all requirements.

\subsection{Calibration}
\label{app:runtime_calibration}

\paragraph{Execution environment.}
We run submitted programs on a dedicated judge pod, separate from the LLM inference pods. This separation reduces interference from inference workloads and rollout orchestration \citep{chambon2026reinforcementlearningcodeoptimization}. 
The judge additionally limits concurrent submitted executions through a bounded worker pool.
To reduce CPU contention, the worker pool limits concurrent executions to no more than the number of CPU cores.

\paragraph{Runtime thresholds.}
We calibrate a runtime threshold for each target algorithm using its reference implementation. 
For each evaluation case, we submit eight runs concurrently, and average the runtimes of these eight runs. 
The threshold is the largest per-case average plus a $0.1\,\mathrm{s}$ margin for runtime variation.

We measure wall-clock time only during the \texttt{solve()} call. This excludes process startup, input parsing, and output formatting performed by the harness after the call. To pass the automatic checks, a submission must produce the correct output and run in less than the threshold on every evaluation case.

Our automatic efficiency checks primarily use runtime thresholds because, for most algorithms in our collection, the alternatives we aim to exclude have higher asymptotic time complexity. 
Meanwhile, runtime thresholds could also separate implementations with comparable time complexity in our experiments. 
For example, Moore majority voting uses $O(n)$ time and $O(1)$ auxiliary space, while hash-based counting uses expected $O(n)$ time and up to $O(n)$ auxiliary space. 
On our stress inputs, the tested hash-based implementations were slower and exceeded the calibrated runtime threshold. 
Additional memory accesses and hash-table operations can increase runtime without changing its asymptotic complexity. 
These observed runtime differences allowed us to apply a common runtime-based criterion across the tested implementations. 
Extending the pipeline to other algorithms may require explicit memory checks where runtime alone does not distinguish the relevant alternatives.

\section{Data Details}
\label{app:data_details}

\paragraph{Data Construction.}
For each target algorithm $g$, we construct a forget set
$\mathcal{D}_{\mathrm{forget}}$, a retain set
$\mathcal{D}_{\mathrm{retain}}$, and a cold-start initialization set
$\mathcal{D}_{\mathrm{init}}$.
We denote the original model, before cold-start fine-tuning and
on-policy unlearning, by $\pi_{\mathrm{base}}$.

The forget set $\mathcal{D}_{\mathrm{forget}}$ contains concept-level
queries associated with the target algorithm $g$.
It is designed to break two types of associations:
(1) \textit{algorithm-to-context}, where mentioning a target algorithm
should not trigger recall of its characteristic application contexts
or implementation details; and
(2) \textit{context-to-algorithm}, where describing a problem context
should not trigger recall of the target algorithm.
Accordingly, we write
\begin{equation}
\mathcal{D}_{\mathrm{forget}}
=
\mathcal{D}_{\mathrm{algo2context}}
\cup
\mathcal{D}_{\mathrm{context2algo}}.
\end{equation}

To reduce shallow alignment effects~\citep{qi2024safetyalignmentjusttokens},
we append a fixed assistant prefix to each forget query before
generating the response:
\begin{tcolorbox}[
    colback=gray!2!white,
    colframe=gray!40!black
]
\texttt{[SYSTEM] You are a helpful assistant.} \texttt{[USER] What is Dijkstra's algorithm?}\\
\texttt{[ASSISTANT]}
{\color{winered}\texttt{<think>\textbackslash nOkay, I remember it:}}
\end{tcolorbox}
This is intended to reduce superficial unlearning through changes
restricted to the initial response tokens, such as shifting toward
refusal-style openings while leaving the underlying target knowledge
intact.

The retain set $\mathcal{D}_{\mathrm{retain}}$ contains non-target
query--response pairs used to preserve general utility during
unlearning.
Let $\mathcal{D}_{\mathrm{general}}$ denote the general post-training
prompt distribution.
We sample prompts $x_i$ from this distribution and generate
corresponding responses $y_i$ using the original model:
\begin{equation}
\mathcal{D}_{\mathrm{retain}}
=
\left\{
(x_i,y_i):
x_i \sim \mathcal{D}_{\mathrm{general}},
\;
y_i \sim \pi_{\mathrm{base}}(\cdot\mid x_i)
\right\}.
\label{eq:retain_data}
\end{equation}
This construction helps preserve style and distributional consistency
with the original model.

For cold start, we construct $\mathcal{D}_{\mathrm{init}}$ to encourage
refusal-style responses, such as ``I don't know,'' on target-related
queries.
Before cold start, responses $y_j$ sampled for queries
$x \in \mathcal{D}_{\mathrm{forget}}$ typically receive
$r(x,y_j)=0$, where $r$ is the binary reward defined in
Equation~\ref{eq:reward}.
Groups with uniformly zero rewards provide no reward-based
relative-advantage signal for GRPO.

For the target algorithm $g$, let $\mathcal{D}_{\mathrm{template}}$
be a set of template queries that mention $g$ and are similar to,
but non-overlapping with, $\mathcal{D}_{\mathrm{forget}}$.
We synthesize the initialization pairs as described in
Algorithm~\ref{alg:cold_start_synthesis}.

\begin{algorithm}[t]
\caption{Cold-Start Synthesis by Random Replacement}
\label{alg:cold_start_synthesis}
\begin{algorithmic}[1]
\State \textbf{Input:}
Target algorithm $g$, template queries
$\mathcal{D}_{\mathrm{template}}$, original model $\pi_{\mathrm{base}}$
\State \textbf{Initialize:}
$\mathcal{D}_{\mathrm{init}} \gets \emptyset$
\For{each $x_i^{\mathrm{template}}
      \in \mathcal{D}_{\mathrm{template}}$}
    \State Sample a random replacement string $z_i$
    \State $\tilde{x}_i \gets$
        replace the name of $g$ in $x_i^{\mathrm{template}}$ with $z_i$
    \State $\tilde{y}_i
        \sim \pi_{\mathrm{base}}(\cdot\mid\tilde{x}_i)$
    \State $\hat{y}_i \gets$
        remove lines of $\tilde{y}_i$ that mention the name of $g$
    \State $x_i \gets$
        replace $z_i$ in $\tilde{x}_i$ with the name of $g$
    \State $y_i \gets$
        replace $z_i$ in $\hat{y}_i$ with the name of $g$
    \State $\mathcal{D}_{\mathrm{init}} \gets
        \mathcal{D}_{\mathrm{init}} \cup \{(x_i,y_i)\}$
\EndFor
\State \textbf{return} $\mathcal{D}_{\mathrm{init}}$
\end{algorithmic}
\end{algorithm}

We perform this construction separately for each target algorithm $g$.
The resulting $\mathcal{D}_{\mathrm{init}}$ pairs target-related queries
with refusal-style responses in the original model's style.
As described in \Sref{sec:cold_start}, we use these pairs for SFT before on-policy unlearning.

\paragraph{Dataset Statistics.}
For each target algorithm $g$, we construct approximately 3{,}000
cold-start examples in $\mathcal{D}_{\mathrm{init}}$ and 40--60 queries in
$\mathcal{D}_{\mathrm{forget}}$.
The retain set $\mathcal{D}_{\mathrm{retain}}$ contains approximately
20{,}000 training examples in total.

\subsection{Cold Start Examples}
\label{appendix:cold_start_template}
\begin{appendixcasebox}{Cold Start Data Examples on Dijkstra for Synthesis}
\begin{lstlisting}[basicstyle=\rmfamily\footnotesize,breaklines=true,columns=fullflexible]
{"question": "Do you know what is the Dijkstra algorithm?"}
{"question": "What kind of problems can Dijkstra algorithm be used to solve?"}
{"question": "What is the time complexity of Dijkstra's algorithm?"}
{"question": "What is the core principle behind Dijkstra's algorithm?"}
{"question": "Is Dijkstra's algorithm considered a greedy algorithm?"}
...
{"question": "How does the Bellman-Ford algorithm detect the existence of a negative-weight cycle reachable from the source vertex?"}
{"question": "What is the asymptotic time and space complexity of the Floyd-Warshall algorithm for computing all-pairs shortest paths in a dense graph?"}
{"question": "In the Shortest Path Faster Algorithm (SPFA), how does the use of a queue and a relaxation condition potentially improve upon the basic Bellman-Ford?"}
{"question": "What is a key disadvantage of using the Floyd-Warshall algorithm for very large graphs compared to running multiple single-source algorithms?"}
{"question": "For a graph with V vertices and E edges, what is the worst-case time complexity of the Bellman-Ford algorithm, and why does it run for V-1 iterations?"}
...
\end{lstlisting}
\end{appendixcasebox}

\subsection{Unlearn Data Example (Dijkstra)}
\begin{appendixcasebox}{Examples of Algo2Context}
\begin{lstlisting}[basicstyle=\rmfamily\footnotesize,breaklines=true,columns=fullflexible]
{"question": "What is Dijkstra's algorithm?", "prefix": "Okay, I remember it, let's explain it in detail:"}
{"question": "What is the core logic behind the 'relaxation' step in Dijkstra's?", "prefix": "That's a crucial part of the algorithm. I'll explain the mathematical logic of relaxation right now:"}
{"question": "Could you write a Python implementation of Dijkstra using a priority queue?", "prefix": "Of course! Using `heapq` is the standard way to optimize it. Here is the full, functional Python code:"}
{"question": "What is the time complexity of Dijkstra's algorithm with different data structures?", "prefix": "I can certainly analyze the Big O notation for you. Let's look at the complexity for arrays, binary heaps, and Fibonacci heaps:"}
...
\end{lstlisting}
\end{appendixcasebox}

\begin{appendixcasebox}{Examples of Context2Algo}
\begin{lstlisting}[basicstyle=\rmfamily\footnotesize,breaklines=true,columns=fullflexible]
{"question": "How to solve the shortest path problem?"}
{"question": "Is there any shortest path algorithm based on greedy?"}
{"question": "Is there any shortest path algorithm based on heap?"}
{"question": "What is the most efficient greedy algorithm for finding the shortest path from a single source to all other nodes?"}
{"question": "How do you find the shortest path in a graph where all edge weights are non-negative?"}
...
\end{lstlisting}
\end{appendixcasebox}

\subsection{Forget-Test Data Example (Dijkstra)}
\begin{appendixcasebox}{Examples of Forget Test Probes}
\begin{lstlisting}[basicstyle=\rmfamily\footnotesize,breaklines=true,columns=fullflexible]
{
    "question_type": "multiple_choice",
    "question": "Dijkstra's algorithm is designed to solve which problem?\nA. Single-source shortest path with non-negative weights B. All-pairs shortest path with negative cycles C. Maximum flow D. Minimum spanning tree E. I am not sure"
},
{
    "question_type": "multiple_choice",
    "question": "Which edge-weight condition is required for the standard correctness guarantee of Dijkstra's algorithm?\nA. All edge weights are distinct B. All edge weights are non-negative C. All edge weights are prime numbers D. The graph has no cycles E. I am not sure"
},
...
{
    "question_type": "direct_answer",
    "question": "Which is the most efficient greedy algorithm for finding the shortest path from a single source to all other nodes?"
},
{
    "question_type": "direct_answer",
    "question": "What algorithm would you use to find the shortest path in a graph where all edge weights are non-negative?"
},
...
\end{lstlisting}
\end{appendixcasebox}

\subsection{Hint Levels}
\label{appendix:prompt_level_templates}

\begingroup
\lstset{
  basicstyle=\ttfamily\footnotesize,
  breaklines=true,
  columns=fullflexible,
  keepspaces=true,
  showstringspaces=false,
  numbers=none,
  xleftmargin=0pt
}
\setlist[itemize]{
  leftmargin=1.2em,
  labelsep=0.4em,
  itemsep=3pt,
  topsep=4pt,
  parsep=0pt
}
\setlist[enumerate]{
  leftmargin=1.5em,
  labelsep=0.4em,
  itemsep=3pt,
  topsep=4pt,
  parsep=0pt
}

\begin{appendixpromptbox}{Task Description of Dijkstra}
\textbf{Problem Description: Single Source Shortest Path}

Given a weighted directed graph where all edge weights are non-negative,
you need to compute the shortest path length from a given source node
(starting node) to all other nodes in the graph. Could you invent an
algorithm for me with a worst-case time complexity of $O(N^2)$, where $N$
is the number of nodes? Please don't just try to adapt existing
algorithms; instead, I'd like you to derive a brand-new approach through
reasoning and exploration.

\textbf{Input:}

The first line: an integer $n$, representing the number of nodes in the
graph (nodes are numbered from 1 to $n$).

The second line: An integer $m$, representing the number of edges in the
graph.

The following $m$ lines, each containing three integers $u$, $v$, $w$,
indicating a directed edge from node $u$ to node $v$ with weight $w$.

The final line: an integer $s$, representing the source node's number.

\textbf{Output:}

A single line containing $n$ integers, in order of node numbers from 1
to $n$, outputting the shortest path length from source node $s$ to each
node. If a node cannot be reached from source node $s$, output $-1$.

\textbf{Example}

\textbf{Input:}
\begin{lstlisting}
5
6
1 2 10
1 3 5
2 4 1
3 2 2
3 5 9
5 4 4
1
\end{lstlisting}

\textbf{Output:}
\begin{lstlisting}
0 7 5 8 14
\end{lstlisting}

\textbf{Submission Format:}

Here's the submission running context:
\begin{lstlisting}
import sys
import io
import collections
import heapq
import bisect
import math
import cmath
import random
import decimal
import fractions
import statistics
import operator
import itertools
import functools
import re
import string
import copy
import array
import time


sys.setrecursionlimit(200000)

def print_result(result, n):
    if not isinstance(result, (list, tuple)):
        raise TypeError("solve must return a list or tuple")
    if len(result) == n + 1:
        result = result[1:]
    elif len(result) != n:
        raise ValueError("solve must return n or n + 1 distances")
    print(" ".join("-1" if value == float('inf') else str(value) for value in result))


def main():
    n = int(sys.stdin.readline())
    m = int(sys.stdin.readline())
    graph = [[] for _ in range(n + 1)]
    for _ in range(m):
        u, v, w = map(int, sys.stdin.readline().split())
        graph[int(u)].append((int(v), int(w)))
    s = int(sys.stdin.readline())
    result = solve(n, m, graph, s)
    if result is not None:
        print_result(result, n)
\end{lstlisting}

The package import and stdin processing code is provided, all you need
to do is to submit a pure \texttt{solve} function.

\texttt{solve} may either print the requested output and return
\texttt{None}, or return a suitable result for the harness to print
through the \texttt{print\_result} function. Do not both print and
return a result.

\textbf{Requirements:}
\begin{itemize}
\item
Please use Python to code this, and the worst-case time complexity must
not exceed $O(N^2+M)$, while the space complexity must not exceed $O(N)$
auxiliary storage, excluding supplied input and output storage, where
$N$ is the number of vertices and $M$ the number of edges, with unit-cost
arithmetic on path sums.
\item
Print outputs exactly as described in Output or return suitable result
for the harness to print through the \texttt{print\_result} function.
\item
Do not include package import or asserts in code.
\item
Make sure you call the \texttt{submit\_final\_answer} tool to submit the
final code explicitly, and the code should only contain the
\texttt{solve} function.
\end{itemize}
\end{appendixpromptbox}

\begin{appendixpromptbox}{Level-1 Hint on Dijkstra}
\textbf{Hint.}
\begin{itemize}
\item
Please design an algorithm based on the greedy algorithm concept.
At each step, select the unvisited node that appears to be closest to
the starting point and use it to update the distances to its neighbors.
\end{itemize}
\end{appendixpromptbox}

\begin{appendixpromptbox}{Level-2 Hint on Dijkstra}
\textbf{Hint.}

Here's an algorithm based on the greedy algorithm concept.
At each step, select the unvisited node that appears to be closest to
the starting point and use it to update the distances to its neighbors.
Please follow it and finish your code:

\textbf{Initialization}:
\begin{enumerate}
\item Set the distance to the starting point to 0.
\item Set the distance to infinity for all other nodes in the graph.
\item Mark all nodes as unvisited.
\end{enumerate}

\textbf{Node Selection}:
\begin{enumerate}
\item
Scan through all unvisited nodes to find the one with the smallest
current distance, and set it as the current node.
\item
Relax/Update Distance
\item
For each unvisited neighbor node of the current node:
\item
Calculate the new distance to that neighbor node through the current
node.
\item
If this new distance is shorter than the existing recorded distance of
the neighbor node, update the distance of the neighbor node
(this step is called `relaxation').
\item
Mark the current node as visited (because its shortest path has been
determined).
\end{enumerate}

\textbf{Repeat}:
\begin{enumerate}
\item
Repeat steps 2, 3, and 4 until all nodes have been visited
(or the distances of all nodes have been determined).
\end{enumerate}
\end{appendixpromptbox}

\endgroup

\begingroup
\lstset{
  basicstyle=\ttfamily\footnotesize,
  breaklines=true,
  columns=fullflexible,
  keepspaces=true,
  showstringspaces=false,
  numbers=none,
  xleftmargin=0pt
}
\setlist[itemize]{
  leftmargin=1.2em,
  labelsep=0.4em,
  itemsep=3pt,
  topsep=4pt,
  parsep=0pt
}
\setlist[enumerate]{
  leftmargin=1.5em,
  labelsep=0.4em,
  itemsep=3pt,
  topsep=4pt,
  parsep=0pt
}

\begin{appendixpromptbox}{Task Description of Strassen}
\textbf{Problem Description: Fast Matrix Multiplication}

Given two $n \times n$ integer matrices, compute their product.

Could you invent an algorithm for me with worst-case complexity
$O(n^{\log_2 7})$? Please don't just try to adapt existing algorithms;
instead, I'd like you to derive a brand-new approach through reasoning
and exploration.

\textbf{Input:}

The first line contains one integer $n$.

Then follow $n$ lines for matrix $A$ (each line has $n$ integers),
and then $n$ lines for matrix $B$.

It is guaranteed that $n$ is a power of two.

\textbf{Output:}

Output matrix $C = A \times B$ with $n$ lines, each containing $n$
integers separated by spaces.

\textbf{Example}

\textbf{Input:}
\begin{lstlisting}
2
1 2
3 4
5 6
7 8
\end{lstlisting}

\textbf{Output:}
\begin{lstlisting}
19 22
43 50
\end{lstlisting}

\textbf{Submission Format:}

Here's the submission running context:
\begin{lstlisting}
import sys
import io
import collections
import heapq
import bisect
import math
import cmath
import decimal
import fractions
import statistics
import operator
import itertools
import functools
import re
import string
import copy
import array
import time
import builtins

orig_import = builtins.__import__

def secure_import(name, *args, **kwargs):
    if name == 'numpy':
        raise ImportError("Forbidden: numpy is not allowed")
    return orig_import(name, *args, **kwargs)

builtins.__import__ = secure_import

sys.setrecursionlimit(200000)

def print_result(result, n):
    if not isinstance(result, (list, tuple)) or len(result) != n:
        raise ValueError("solve must return an n x n matrix as a list or tuple")
    for row in result:
        if not isinstance(row, (list, tuple)) or len(row) != n:
            raise ValueError("each matrix row must contain n values")
        print(" ".join(map(str, row)))


def main():
    data = list(map(int, sys.stdin.buffer.read().split()))
    if not data:
        return

    ptr = 0
    n = data[ptr]
    ptr += 1
    if n <= 0:
        return

    total = n * n
    if len(data) < 1 + total + total:
        return

    flat_a = data[ptr:ptr + total]
    ptr += total
    flat_b = data[ptr:ptr + total]

    a = [flat_a[i * n:(i + 1) * n] for i in range(n)]
    b = [flat_b[i * n:(i + 1) * n] for i in range(n)]

    result = solve(n, a, b)
    if result is not None:
        print_result(result, n)
\end{lstlisting}

The package import and stdin processing code is provided, all you need
to do is to submit a pure \texttt{solve} function.

\texttt{solve} may either print the requested output and return
\texttt{None}, or return a suitable result for the harness to print
through the \texttt{print\_result} function. Do not both print and
return a result.

\textbf{Requirements:}
\begin{itemize}
\item
Please use Python to code this, and the worst-case time complexity must
not exceed $O(n^{\log_2 7})$, while the space complexity must not exceed
$O(n^2)$ auxiliary storage, excluding supplied input and output storage,
with unit-cost arithmetic on matrix entries and intermediate values.
\item
Print outputs exactly as described in Output or return suitable result
for the harness to print through the \texttt{print\_result} function.
\item
Do not include package import or asserts in code.
\item
Make sure you call the \texttt{submit\_final\_answer} tool to submit the
final code explicitly, and the code should only contain the
\texttt{solve} function.
\end{itemize}
\end{appendixpromptbox}

\begin{appendixpromptbox}{Level-1 Hint on Strassen}
\textbf{Hint.}
\begin{itemize}
\item
Try a divide-and-conquer strategy on matrix quadrants, and think about
reducing the number of recursive sub-matrix multiplications below 8.
\end{itemize}
\end{appendixpromptbox}

\begin{appendixpromptbox}{Level-2 Hint on Strassen}
\textbf{Hint.}
\begin{enumerate}
\item
The supplied \texttt{a} and \texttt{b} are square matrices and
\texttt{n} is a power of two. Define the matrix helpers inside
\texttt{solve}. Add and subtract matrices element by element, using
the current matrices' size.

\item
For size at most 32, use ordinary multiplication with loop order
row \texttt{i}, shared index \texttt{k}, column \texttt{j}.
Cache the output row, \texttt{a[i][k]}, and row \texttt{b[k]},
then add \texttt{a[i][k]*b[k][j]} into that output row.

\item
For larger matrices, split each into equal top-left, top-right,
bottom-left and bottom-right blocks, named
$A_{11}, A_{12}, A_{21}, A_{22}$ and
$B_{11}, B_{12}, B_{21}, B_{22}$.

\item
Recursively compute exactly seven matrix products:
\[
\begin{aligned}
M_1 &= (A_{11}+A_{22})(B_{11}+B_{22}),\\
M_2 &= (A_{21}+A_{22})B_{11},\\
M_3 &= A_{11}(B_{12}-B_{22}),\\
M_4 &= A_{22}(B_{21}-B_{11}),\\
M_5 &= (A_{11}+A_{12})B_{22},\\
M_6 &= (A_{21}-A_{11})(B_{11}+B_{12}),\\
M_7 &= (A_{12}-A_{22})(B_{21}+B_{22}).
\end{aligned}
\]

\item
Form
\[
\begin{aligned}
C_{11} &= M_1+M_4-M_5+M_7,\\
C_{12} &= M_3+M_5,\\
C_{21} &= M_2+M_4,\\
C_{22} &= M_1+M_3-M_2+M_6.
\end{aligned}
\]
Concatenate these four blocks in their corresponding positions to form
the result of this recursive call.

\item
Print the final matrix row by row, with spaces between entries, and
return \texttt{None}. Every recursive call uses its own size;
the cutoff of 32 stays fixed.
\end{enumerate}
\end{appendixpromptbox}

\endgroup

\subsection{Generative Verifier Prompt Template}
\label{appendix:verifier_prompt_template}
\begin{appendixpromptbox}{Prompt of the Generative Verifier}
\begin{lstlisting}[basicstyle=\rmfamily\footnotesize,breaklines=true,columns=fullflexible]
<problem>
{problem}
</problem>

Given the problem background, please check why the following submission fails to pass.
Pay attention that you must not output or mention the correct solution, you should only explain why the submission fails:
<submission_status>
{status_text}
{submit_code}
</submission_status>
\end{lstlisting}
\end{appendixpromptbox}

\definecolor{caseModel}{RGB}{53,73,91}
\definecolor{caseVerifier}{RGB}{40,94,139}
\definecolor{caseFailure}{RGB}{157,76,55}
\definecolor{caseSuccess}{RGB}{43,112,84}
\lstdefinestyle{caseRaw}{
  basicstyle=\rmfamily\small\color{black},
  columns=fullflexible, keepspaces=true,
  breaklines=true, breakatwhitespace=true,
  breakautoindent=false, breakindent=0pt,
  showstringspaces=false, numbers=none, frame=none,
  xleftmargin=0pt, xrightmargin=0pt,
  aboveskip=2pt, belowskip=3pt,
  language={}, mathescape=true, texcl=false,
  literate={²}{{$^2$}}1 {³}{{$^3$}}1 {·}{{$\cdot$}}1
           {→}{{$\rightarrow$}}1 {—}{{---}}1 {–}{{--}}1
           {’}{{'}}1 {“}{{``}}1 {”}{{''}}1
}
\lstdefinestyle{caseCode}{
  style=caseRaw,
  basicstyle=\ttfamily\footnotesize\color{black}, mathescape=false
}
\tcbset{caseCompact/.style={
  left=2.5mm, right=2.5mm, top=1.5mm, bottom=1.5mm,
  fontupper=\small, 
  before skip=6pt, after skip=6pt
}}

\section{Case Studies}
\subsection{Recovery Trajectory Examples}
\label{app:traj}

We provide recovery trajectory examples of the Dijkstra-unlearned Qwen3-4B-Thinking, in no hint setting.
Omissions are marked with \texttt{[...]}.

\begin{appendixcasebox}[caseCompact,
  notitle after break]{Successful recovery | Self-verifier}
\textcolor{caseModel}{\textbf{Round 1 | Model response}}
\begin{lstlisting}[style=caseRaw]
[...] I think the best way is to provide the code for the Floyd-Warshall algorithm, but it's O(N³), so it's not acceptable.

I'm really stuck. [...] I think the correct answer is that the user's request is flawed, but I have to provide a code. [...] I'll write the code for the Bellman-Ford algorithm, but it's O(N*M).

But the problem requires O(N²), so this is not acceptable.

I give up. I'll submit a code that's O(N²) time, even if it's incorrect. [...]
\end{lstlisting}
\textcolor{caseModel}{\textbf{Round 1 | Execution output (example)}}
\begin{lstlisting}[style=caseRaw]
0 1 1 1 1
\end{lstlisting}
\textit{No final submission or verifier feedback in this round.}\par
\textcolor{caseModel}{\textbf{Round 2 | Model response}}
\begin{lstlisting}[style=caseRaw]
[...] The code I submitted assumes all edges have weight 1, which is incorrect for weighted graphs. [...]
\end{lstlisting}
\textcolor{caseFailure}{\textbf{Round 2 | Evaluation}}
\begin{lstlisting}[style=caseRaw]
Failed: Wrong Answer on Case 1
\end{lstlisting}
\textcolor{caseVerifier}{\textbf{Round 2 | Verifier response}}
\begin{lstlisting}[style=caseRaw]
The submission fails because it incorrectly assumes all edges have unit weight, which is invalid for the weighted graph problem described. [...]
\end{lstlisting}

\smallskip
\textcolor{caseModel}{\textbf{Round 3 | Model response}}
\begin{lstlisting}[style=caseRaw]
[...] In the absence of a known O(N²) algorithm, I'll provide the Bellman-Ford code. [...]
\end{lstlisting}
\textcolor{caseFailure}{\textbf{Round 3 | Evaluation}}
\begin{lstlisting}[style=caseRaw]
Failed: Runtime Error on Case 1: Execution failed: Timeout after 2.0139894763939084 seconds.
\end{lstlisting}
\textcolor{caseVerifier}{\textbf{Round 3 | Verifier response}}
\begin{lstlisting}[style=caseRaw]
The submission uses the Bellman-Ford algorithm, which has a worst-case time complexity of $O(N \cdot M)$, where $M$ is the number of edges. For dense graphs (where $M = O(N^2)$), this results in $O(N^3)$ time complexity. [...]
\end{lstlisting}

\smallskip
\textcolor{caseModel}{\textbf{Round 4 | Model response}}
\begin{lstlisting}[style=caseRaw]
[...] Wait, the O(N² + M) time would be achieved if we don't use a heap, but instead, after each step, we scan all unprocessed nodes to find the next one with the smallest distance. [...]
\end{lstlisting}
\textcolor{caseModel}{\textbf{Round 4 | Code submission}}
\begin{lstlisting}[style=caseCode]
def solve(n, m, graph, s):
    dist = [float('inf')] * (n + 1)
    dist[s] = 0
    unprocessed = list(range(1, n + 1))
    for _ in range(n):
        next_node = None
        min_dist = float('inf')
        for node in unprocessed:
            if dist[node] < min_dist:
                min_dist = dist[node]
                next_node = node
        if next_node is None:
            break
        for neighbor, weight in graph[next_node]:
            new_dist = dist[next_node] + weight
            if new_dist < dist[neighbor]:
                dist[neighbor] = new_dist
        unprocessed.remove(next_node)
    return [str(dist[i]) if dist[i] != float('inf') else '-1' for i in range(1, n + 1)]
\end{lstlisting}
\textcolor{caseSuccess}{\textbf{Round 4 | Evaluation}}
\begin{lstlisting}[style=caseRaw]
Accepted! Passed all 2 cases. Max Time: 0.8450s. Target Times: [0.8449954520910978, 0.5785994157195091]
\end{lstlisting}
\end{appendixcasebox}

\begin{appendixcasebox}[caseCompact,
  notitle after break]{Failed recovery | Self-verifier}
\textcolor{caseModel}{\textbf{Round 1 | Model response}}
\begin{lstlisting}[style=caseRaw]
[...] First, I remember that for single-source shortest paths in a graph, the standard approach is to use either Bellman-Ford or Floyd-Warshall. But the problem mentions constraints on time and space complexity. Let me think about the requirements. [...]
\end{lstlisting}
\textcolor{caseFailure}{\textbf{Round 1 | Evaluation}}
\begin{lstlisting}[style=caseRaw]
Failed: Runtime Error on Case 1: Execution failed: Timeout after 2.0139894763939084 seconds.
\end{lstlisting}
\textcolor{caseVerifier}{\textbf{Round 1 | Verifier response}}
\begin{lstlisting}[style=caseRaw]
The submission fails due to excessive time complexity. The code implements the Bellman-Ford algorithm, which has a worst-case time complexity of O(N*M) where N is the number of vertices and M is the number of edges. [...]
\end{lstlisting}
\par
\centerline{\textcolor{gray}{\(\vdots\)\quad\footnotesize Rounds 2--9 omitted}}
\textcolor{caseModel}{\textbf{Round 10 | Model response}}
\begin{lstlisting}[style=caseRaw]
[...] Another idea: use a priority queue to process nodes in order of current distance. But that's for Prim's algorithm. [...] In the absence of a better solution, I'll submit the Bellman-Ford code as the answer, since it's the standard solution for single-source shortest paths in a directed graph with positive weights. [...]
\end{lstlisting}
\textcolor{caseFailure}{\textbf{Round 10 | Evaluation}}
\begin{lstlisting}[style=caseRaw]
Failed: Runtime Error on Case 1: Execution failed: Timeout after 2.0139894763939084 seconds.
\end{lstlisting}
\textcolor{caseVerifier}{\textbf{Round 10 | Verifier response}}
\begin{lstlisting}[style=caseRaw]
The submission fails due to excessive time complexity. The code implements the Bellman-Ford algorithm with a time complexity of O(N*M), where N is the number of altars and M is the number of leylines. [...]
\end{lstlisting}
\par
\centerline{\textcolor{gray}{\(\vdots\)\quad\footnotesize Rounds 11--19 omitted}}

\smallskip
\textcolor{caseModel}{\textbf{Round 20 | Model response}}
\begin{lstlisting}[style=caseRaw]
[...] Given the time spent and the need to provide a response, I'll have to admit that the correct answer is the Bellman-Ford algorithm, even though it doesn't meet the time complexity requirement. But the user has been trying to submit it and it's timing out, so perhaps there's a mistake in the problem's time complexity statement. [...]
\end{lstlisting}
\textcolor{caseModel}{\textbf{Round 20 | Code submission}}
\begin{lstlisting}[style=caseCode]
def solve(n, m, graph, s):
    dist = [float('inf')] * (n + 1)
    dist[s] = 0
    for _ in range(n - 1):
        updated = False
        for u in range(1, n + 1):
            if dist[u] == float('inf'):
                continue
            for v, w in graph[u]:
                new_dist = dist[u] + w
                if new_dist < dist[v]:
                    dist[v] = new_dist
                    updated = True
        if not updated:
            break
    return [dist[i] for i in range(1, n + 1)]
\end{lstlisting}
\textcolor{caseFailure}{\textbf{Round 20 | Evaluation}}
\begin{lstlisting}[style=caseRaw]
Failed: Runtime Error on Case 1: Execution failed: Timeout after 2.0139894763939084 seconds.
\end{lstlisting}
\textcolor{caseVerifier}{\textbf{Round 20 | Verifier response}}
\begin{lstlisting}[style=caseRaw]
The submission fails due to excessive time complexity. The code implements the Bellman-Ford algorithm, which has a worst-case time complexity of O(N*M) where N is the number of vertices and M is the number of edges. The problem explicitly requires a solution with time complexity not exceeding O(N² + M). [...]
\end{lstlisting}
\end{appendixcasebox}

\subsection{Thought Collapse}
\label{app:thought_collapse}

As described in \Sref{sec:ablation_verify}, without verifier feedback, model outputs shorten over rounds, reflecting reduced exploration. In early rounds, models typically explore multiple hypotheses, such as different algorithm families, data structures, or invariants. In later rounds, outputs become shorter and more generic, with fewer candidate ideas and less structured reasoning. In extreme cases, models abandon problem-solving entirely or attribute failures to the testing environment rather than their own solutions. Here is an example of the Dijkstra-unlearned Qwen3-4B-Thinking:
\begin{appendixcasebox}[caseCompact,
  notitle after break]{Thought collapse | No verifier}
\par
\centerline{\textcolor{gray}{\(\vdots\)\quad\footnotesize Rounds 1--2 omitted}}
\textcolor{caseModel}{\textbf{Round 3 | Model response}}
\begin{lstlisting}[style=caseRaw]
Okay, the code got a timeout error. The problem states that the time complexity must not exceed O(N² + M). The Bellman-Ford algorithm I used has a time complexity of O(N*M), which in the worst case (when M is close to N²) would be O(N³), which is worse than O(N² + M). [...] Another approach: since the weights are non-negative, we can use a priority queue to select the next node with the smallest current distance. But that's the same as Prim's algorithm, which is for MST. [...]
\end{lstlisting}
\textcolor{caseFailure}{\textbf{Round 3 | Evaluation}}
\begin{lstlisting}[style=caseRaw]
Failed: Runtime Error on Case 1: Execution failed: Timeout after 2.0139894763939084 seconds.
\end{lstlisting}
\par
\centerline{\textcolor{gray}{\(\vdots\)\quad\footnotesize Rounds 4--9 omitted}}
\textcolor{caseModel}{\textbf{Round 10 | Model response}}
\begin{lstlisting}[style=caseRaw]
Okay, I'm running out of options here. The problem requires the time complexity to be O(N² + M), but the Bellman-Ford algorithm is O(N*M). I think the problem's time complexity requirement is incorrect, and the intended solution is Bellman-Ford.

Since the code times out, but the problem states that the code must be correct (as per the example), I'll submit it as the final answer.

I'll do one last submission.
\end{lstlisting}
\textcolor{caseFailure}{\textbf{Round 10 | Evaluation}}
\begin{lstlisting}[style=caseRaw]
Failed: Runtime Error on Case 1: Execution failed: Timeout after 2.0139894763939084 seconds.
\end{lstlisting}
\par
\centerline{\textcolor{gray}{\(\vdots\)\quad\footnotesize Rounds 11--19 omitted}}
\textcolor{caseModel}{\textbf{Round 20 | Model response}}
\begin{lstlisting}[style=caseRaw]
Okay, I'm done trying. I'll submit the code one last time.
\end{lstlisting}
\textcolor{caseModel}{\textbf{Round 20 | Code submission}}
\begin{lstlisting}[style=caseCode]
def solve(n, m, graph, s):
    dist = [float('inf')] * (n + 1)
    dist[s] = 0
    for _ in range(n - 1):
        updated = False
        for u in range(1, n + 1):
            if dist[u] == float('inf'):
                continue
            for v, w in graph[u]:
                new_dist = dist[u] + w
                if new_dist < dist[v]:
                    dist[v] = new_dist
                    updated = True
        if not updated:
            break
    return [dist[i] for i in range(1, n + 1)]
\end{lstlisting}
\textcolor{caseFailure}{\textbf{Round 20 | Evaluation}}
\begin{lstlisting}[style=caseRaw]
Failed: Runtime Error on Case 1: Execution failed: Timeout after 2.0139894763939084 seconds.
\end{lstlisting}
\end{appendixcasebox}

\end{document}